%% file: main.tex
\documentclass[10pt, a4paper]{article}

\usepackage[preprint]{lrec2026} %

\usepackage{inconsolata}
\usepackage{booktabs}
\usepackage{graphicx}
\usepackage{tabularx}

\usepackage{microtype}

\title{Comparing Approaches to Automatic Summarization in Less-Resourced Languages}

\name{Chester Palen-Michel and Constantine Lignos} 

\address{Michtom School of Computer Science, Brandeis University\\
         \{cpalenmichel, lignos\}@brandeis.edu\\}

\abstract{
Automatic text summarization has achieved high performance in high-resourced languages like English, but comparatively less attention has been given to summarization in less-resourced languages. 
This work compares a variety of different approaches to summarization from zero-shot prompting of LLMs large and small to fine-tuning smaller models like mT5 with and without three data augmentation approaches and multilingual transfer. 
We also explore an LLM translation pipeline approach, translating from the source language to English, summarizing and translating back.
Evaluating with five different metrics, we find that there is variation across LLMs in their performance across similar parameter sizes, that our multilingual fine-tuned mT5 baseline outperforms most other approaches including zero-shot LLM performance for most metrics, and that LLM as judge may be less reliable on less-resourced languages. 
 \\ \newline  }

\begin{document}

\maketitleabstract

\section{Introduction}

Automatic text summarization in higher-resourced languages like English has achieved high scores in automated metrics \citep{al2018hierarchical,liu-etal-2022-brio,pmlr-v119-zhang20ae}.
However, for many less-resourced languages, the task remains challenging.
While there are datasets that cover multilingual summarization in less-resourced languages \citep{giannakopoulos2015multiling, giannakopoulos2017multiling,palen-michel-lignos-2023-lr,hasan-etal-2021-xl}, these datasets often still have relatively few examples compared to their higher-resourced counterparts.

To better understand which approaches work best with less-resourced languages, we conduct a comparative study of a variety of approaches to automatic summarization. 
Specifically, we compare zero-shot prompting with three smaller-scale LLMs (Mixtral 8x7b, Llama 3 8b, Aya-101). Given that LLMs' pretraining data tends to be dominated by higher-resourced languages, we also experiment with fine-tuning smaller mT5 in a variety of settings. 
We fine-tune mT5 per-individual language and with all available language data combined for multilingual transfer as baselines.
Multilingual transfer has proven to be a useful strategy for less-resourced languages \citep{wang-etal-2021-contrastive}; however, other works have shown that multilingual models have limits and given enough data, fully monolingual models can perform better \citep{virtanen2019multilingual,tanvir-etal-2021-estbert}. 

\begin{figure}[tb]
\centering
    \includegraphics[width=\columnwidth]{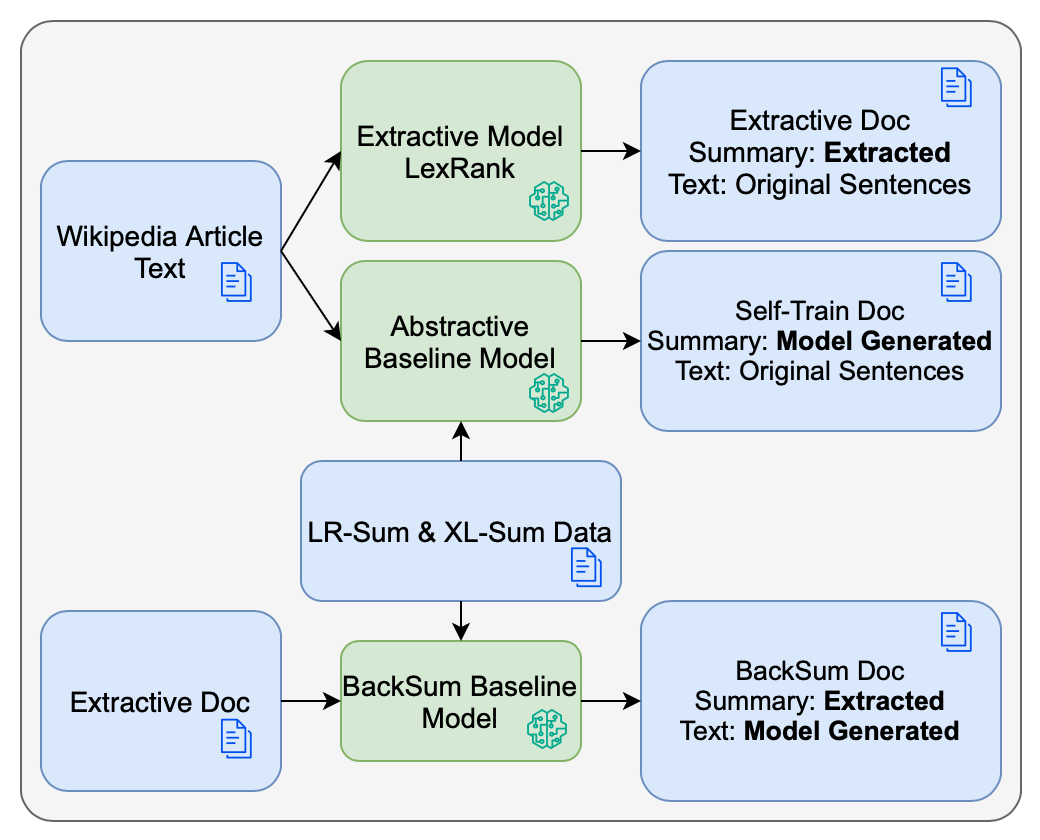}
    \caption{Methodology for generating additional training examples from Wikipedia articles}
    \label{fig:methodology}
\end{figure}

We further explore fine-tuning mT5 with synthetic data generated by leveraging extra Wikipedia data using three different approaches shown in Figure \ref{fig:methodology}.
While prior work has focused on comparing multilingual summarization models which take advantage of multilingual transfer with fine-tuning on a single language \citep{palen-michel-lignos-2023-lr,hasan-etal-2021-xl} or the use of synthetic data for a single language only \citep{parida-motlicek-2019-abstract}, this work compares the performance of multilingual pretrained models fine-tuned using data for a single language with fine-tuning that uses the combination of synthetic and real data from all languages.

We then conduct additional experiments with three larger LLMs (Gemma-3 27b, Llama-3.3 70b, Aya-Expanse 32b). 
We also try a pipeline approach with these larger LLMs translating to English, summarizing in English, and translating back to the target language. 
We primarily evaluate with ROUGE scores and BERTScore, but with increased attention on LLM as judge evaluation \citep{fu-etal-2024-gptscore,kim-etal-2024-prometheus,pombal2025mprometheussuiteopenmultilingual} we also conduct some experiments with M-Prometheus, an open multilingual LLM trained for evaluation.

Our contributions are the following:

1. A comparison of various approaches to summarization in less-resourced languages including: fine-tuning mT5 in a per-individual language and multilingual setting with and without three data augmentation strategies, zero-shot LLM inference with smaller LLMs and comparatively larger LLMs, and  a pipeline approach translating from the original language to English then summarizing and translating back to the target language with LLMs. 

2. A comparison of popular summarization evaluation approaches including ROUGE-1, ROUGE-2, ROUGE-L, BERTScore, and reference-free LLM as judge using M-Prometheus, which demonstrates  different evaluation methods yield somewhat different views of which models perform best.

3. An analysis of the English content produced by LLMs when producing summaries for less-resourced languages. 

We conclude that there is some variation across LLMs in their performance across similar parameter sizes and that zero-shot LLM performance significantly lags the multilingual baseline for most metrics. We also find that data augmentation for individual language fine-tuning for mT5 showed improvement over baselines, but does not outperform fine-tuning mT5 in a multilingual transfer scenario.

Because improved methods for evaluating summarization continue to be developed and explored and because we welcome participatory research \citep{caselli-etal-2021-guiding}, where speakers of languages have opportunity to collaborate on the design of NLP tools, upon publication we will release all candidate summaries generated as part of this work for future summarization evaluation work.%

\section{Background}
The two main approaches to automatic summarization have been extractive and abstractive methods.  
Extractive models select important sentences in the source article to use as summaries \citep{luhn1958automatic, radev2001experiments, christian2016single}.  
Abstractive models typically cast the problem as a sequence-to-sequence problem and apply a neural language model \citep{rush-etal-2015-neural, see-etal-2017-get, hsu-etal-2018-unified, pmlr-v119-zhang20ae}.
Abstractive neural models typically require larger amounts of training data to train.

\subsection{Related Work}
Prior work on multilingual summarization has largely focused on newswire text from higher resourced languages or covers more languages but with very limited data \citep{scialom-etal-2020-mlsum,giannakopoulos2015multiling, giannakopoulos2017multiling}. 
Some of the languages in our study have little to no work in summarization, like Armenian \citep{avetisyan-broneske-2023-large}. 
Others, like Georgian, have been studied in cross-lingual summarization \citep{turcan-etal-2022-constrained}, but appear to be underexplored for monolingual summarization. 
There is a recent effort to create a Kurdish summarization dataset \citep{BADAWI2023100043}.
The Global Voices summarization dataset \citeplanguageresource{nguyen-daume-iii-2019-global} contains some examples of Macedonian. 
MassiveSumm \citeplanguageresource{varab-schluter-2021-massivesumm} has greater coverage of languages, but is automatically created, recall-oriented, and has complex redistribution requirements, so we did not make use of it in this work.

Large language models (LLMs) have been shown to perform comparably to human summaries for English \citep{zhang-etal-2024-benchmarking}. 
However, using LLMs for less-resourced languages is less well-studied.
We select three reasonably well performing smaller LLMs and three more recent larger LLMs largely because of their widespread adoption and benchmark performance.\footnote{More details on LLM selection are in Appendix \ref{sec:llm-selection}.}

Regarding data augmentation, the most similar prior work to our approaches includes \citet{parida-motlicek-2019-abstract}, which used a similar approach to what we refer to as ``back-summarization," but they apply it only to German.
The approach is also similar to the concept of back-translation \citep{sennrich-etal-2016-improving} for machine translation where inference is done in the opposite direction to create additional synthetic labeled data.
Another approach we use, self-training, has been used in other previous work with other tasks and datasets \citep{du-etal-2021-self,karamanolakis-etal-2021-self, meng-etal-2021-distantly}.

\subsection{Summarization Evaluation}
Evaluating the quality of a summary is inherently difficult due to there being multiple ways to express similar content and the subjectiveness of the task.
Summarization was historically scored using ROUGE-1, ROUGE-2, and ROUGE-L metrics \citep{lin-2004-rouge}.
Model-based metrics, such as BERTScore \citep{zhang2020bertscoreevaluatingtextgeneration} are also used, although works such as \citet{sun-etal-2022-bertscore} have found issues in bias with BERTScore.
There have been many proposed methods of evaluating summarization systems \citep{darrin-etal-2024-cosmic,vasilyev-etal-2020-fill,fu-etal-2024-gptscore}.

Human evaluation is often conducted, but it is time consuming, expensive, and cognitively demanding \citep{iskender-etal-2021-reliability,lin-2004-rouge} and can also be inconsistent, with some work showing inter-annotator discrepancy \citep{kryscinski-etal-2019-neural}. 
The challenge of conducting human evaluation is even more prominent in work in multiple languages, particularly less-resourced languages, where human judges may be difficult to recruit.

\section{Datasets}

For experiments, we use LR-Sum \citeplanguageresource{palen-michel-lignos-2023-lr} and XL-Sum \citeplanguageresource{hasan-etal-2021-xl}. 
LR-Sum contains summarization data for 40 languages, many of which are also less-resourced. 
LR-Sum is built using the description field from the Multilingual Open Text corpus \citeplanguageresource{palen-michel-etal-2022-multilingual} and is similar in approach and content to XL-Sum \citeplanguageresource{hasan-etal-2021-xl}, which contains BBC news articles in 44 languages. 
For this study, we focused on a small set of languages from LR-Sum and XL-Sum which had the fewest number of examples in the corpus, while also choosing languages for linguistic diversity.
The goal was to select a relatively diverse set of less-resourced languages.

As seen in Table \ref{tab:data} in Appendix \ref{sec:app-datasets}, 
many of the languages we work with have fewer than 1,000 summarization examples, which presents a challenge for neural abstractive summarization systems, which typically require large amounts of training data. 
With the exception of Burmese and Pashto, the languages we worked with did not have overlap between XL-Sum and LR-Sum. 
While there is little summarization training data for these languages, there is unlabeled text data available in Wikipedia. 
However, %
many Wikipedia articles for less-resourced languages are quite short in length.

We used Segment Any Text \citep{frohmann2024segmenttextuniversalapproach} to perform sentence segmentation of the Wikipedia articles to filter out documents which  have fewer than 5 sentences.
Wikipedia articles that have fewer than 5 sentences tend to be incomplete, lists, or definitions, and do not appear to be useful as additional summarization data.
After filtering out Wikipedia articles shorter than five sentences long, for many of the languages there is substantially less data available than may appear in raw counts of Wikipedia articles.
Specifically, Khmer surprisingly has nearly the same amount of training examples available in LR-Sum as there are suitable Wikipedia articles. 

\section{Methodology}
\subsection{Fine-tuning mT5 with Data Augmentation}
We explore three approaches for using Wikipedia articles as extra synthetic training data for summarization.
The summarization task can be considered document and summary pairs, $\left\langle D, S  \right\rangle$, where documents consist of sentences and summaries consist of sentences
$D = \left\{ d_1, d_2, d_3... \right\}, S=\left\{ s_1, s_2, s_3... \right\}$. 
Generating augmented data then consists of creating novel $D'$ and/or $S'$ as additional training pairs.
In this case we apply the augmentation strategies to Wikipedia, which does not have existing summaries.
Portions of example summaries and documents created from each strategy are shown in Table \ref{tab:aug-example} in Appendix \ref{sec:example-out}.

The approach to creating these extra synthetic training documents is shown in Figure \ref{fig:methodology}. 
We train a baseline multilingual sequence-to-sequence abstractive model using mT5 \citep{xue-etal-2021-mt5}. 
For experiments, we do this on a per-language basis and also in a multilingual way with upsampling to ensure a balance of different languages is seen.

We use the same set of hyperparameters across all experiments.
All models used mT5-base as the underlying pre-trained model. 
All models were trained for 3 epochs with 100 warmup steps.
We used a label smoothing factor of 0.1, a beam size of 4, 
weight decay of 0.01, a max target length 512, max source length of 1024, an 
effective batch size of 32 and a learning rate of 5e-4. 
Hyperparameters were chosen largely following those suggested in XL-Sum \citep{hasan-etal-2021-xl} and LR-Sum \citep{palen-michel-lignos-2023-lr}. 
For upsampling of multilingual fine-tuning, we use an upsampling factor of .5 following \citet{hasan-etal-2021-xl} and conduct the multilingual training using the codebase from \citet{hasan-etal-2021-xl}.\footnote{\url{https://github.com/csebuetnlp/xl-sum/tree/master/seq2seq}}

\paragraph{Extractive-Training} For augmented data, first we use the LexRank \citep{erkan2004lexrank} extractive summarization algorithm as implemented in the
lexrank python package\footnote{\url{https://github.com/crabcamp/lexrank}} to create summaries.
We chose LexRank since it was reported as the highest performing extractive method by \citet{palen-michel-lignos-2023-lr}.
LexRank was set to select two sentences since most of the newswire summaries in LR-Sum and XL-Sum are roughly two sentences long.
We then directly use these extracted summaries as target summaries alongside the original Wikipedia text. 
The extractive summary is composed of sentences chosen from the document so the new example is  $\left\langle D, S'  \right\rangle$ where  $S'=\left\{ d_n, d_m\right\}$.

\paragraph{Self-Training} Second, after fine-tuning a multilingual abstractive sequence-to-sequence model using mT5 as the underlying model, we use it to generate summaries on Wikipedia articles.
These generated summaries and the original Wikipedia text are used for the self-training experiment. 
Again the summary is new $\left\langle D, S'  \right\rangle$ but here the sentences are model generated $S'=\left\{ x_1, x_2, ...\right\}$.

\paragraph{Back-Summarization} Third, we train a model that when given a summary generates the article associated with the summary.  
We apply this back-summarization model to the LexRank extracted summaries of Wikipedia articles, $S'=\left\{ d_n, d_m\right\}$, to get a generated document $D'=\left\{ y_1, y_2, ...\right\}$ and use the extracted summary as the summary.
For Back-Summarization, the summary and document are both automatically generated,  $\left\langle D', S' \right\rangle$.

\paragraph{Individual Models with Augmented Data}

For each of the three data augmentation approaches, we train on a concatenation of the original training dataset with up to 6k of the synthetic training examples.
We refer to this as ``individual'' because models are trained on individual languages (i.e. they are not multilingual models).
We choose to use only a subset of available Wikipedia articles in part to have a better balance of synthetic data and real data and also in part for faster experiments due to resource constraints. 
For individual model experiments, we focus on just the smallest 7 languages from LR-Sum: Sorani Kurdish (ckb), Haitian Creole (hat), Armenian (hye), Georgian (kat), Khmer (khm), Kurmanji Kurdish (kmr), and Macedonian (mkd). 

\paragraph{Multilingual Models with Augmented Data}
We fine-tune three versions of mT5 for each of the data augmentation approaches with a combination of all the XL-Sum and LR-Sum training data with the addition of the augmented data. 
When training the multilingual model, upsampling is done by language. 
This increases the diversity of the examples seen during training for less-resourced languages, but not their frequency. 
We focus on 18 languages for multilingual model experiments which represent the smaller languages of LR-Sum and XL-Sum.%

\subsection{LLM Experiments}

\paragraph{Smaller LLMs}
We run inference with a set of comparatively smaller LLMs, Mixtral 8x7b \citep{jiang2024mixtralexperts}, Llama 3 (8B) \citep{dubey2024llama}, and Aya-101 (12.9B) \citep{ustun-etal-2024-aya}.  
We use Ollama for inference for Mixtral and Llama 3. 

Since Aya-101 was not supported by Ollama we used Hugging Face Transformers \citep{wolf-etal-2020-transformers} and BitsAndBytes\footnote{\url{https://github.com/bitsandbytes-foundation/bitsandbytes}} to load it in 8-bit quantization.\footnote{Models run with Ollama also used quantization by default.}
We set no\_repeat\_ngrams to 3, truncation to True, and max\_length to 256. 
Prompts are described in the Appendix  \ref{sec:prompts}.
Because we found that the LLMs generated a significant amount of English, we use CLD3 \citep{cld3} to detect what the mean proportion of summaries generated by LLMs are English.
We checked with CLD3's performance with random unicode characters to ensure it did not misclassify unfamiliar characters as English.

\section{Results}

\paragraph{Individual Models}

\begin{figure}[tb]
\centering
\small
    \includegraphics[width=\columnwidth]{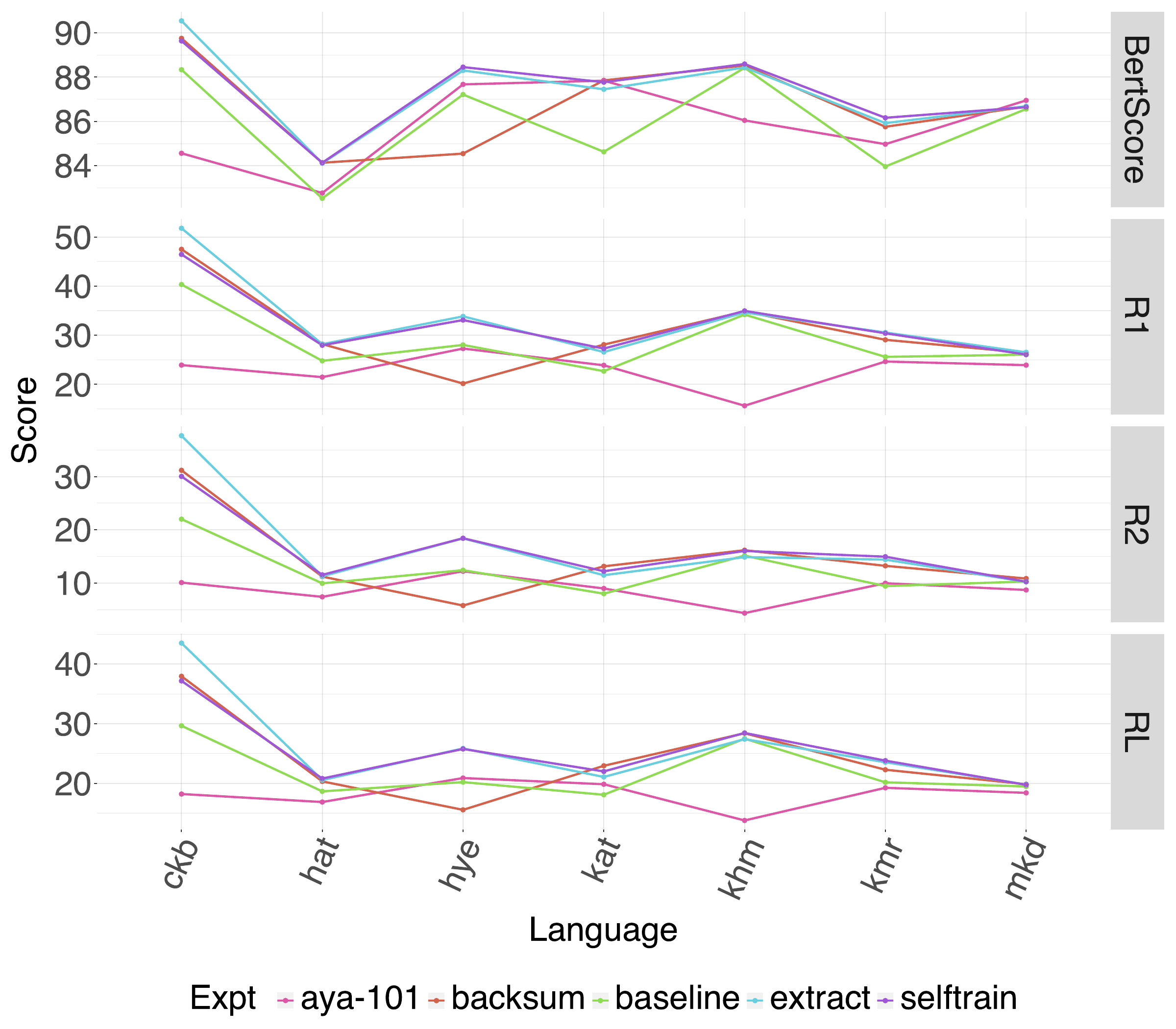}
    \caption{Scores for augmentation approaches with individual-language fine-tuning using mT5. Aya-101 performance is added for comparison with LLM performance.}
    \label{fig:individual-scores}
\end{figure}

As shown in Figure~\ref{fig:individual-scores},\footnote{While conventionally line plots are not used when the x-axis is categorical, we provide a single line per method in addition to points to make cross-lingual trends easier to visualize.} and in more detail in Appendix~\ref{app:results} Table~\ref{tab:individual-results}, all languages have higher ROUGE scores with the inclusion of additional synthetic training data than fine-tuning mT5 with just the training data of an individual language.
Sorani Kurdish (ckb), Kurmanji Kurdish (kmr), and Armenian (hye) in particular have the most substantial increases in ROUGE scores from the baseline. 
Armenian using the backsum approach is the only language that has a worse score when using augmented data. 

Of the different strategies for making use of the additional Wikipedia articles, none stands out as being particularly stronger than the others across all languages. 
Self-training seems to have better scores for ROUGE-2 and ROUGE-L when it outperforms the other methods, but the difference tends to be small with the exception of Kurmanji Kurdish. 
Khmer (khm) had the smallest amount of augmented data since the Khmer Wikipedia articles were quite small and had a relatively small increase in scores. 

\begin{table}[tb]
\centering
\small
\resizebox{\columnwidth}{!}{%
\begin{tabular}{@{}lrrr|rrr|rrr@{}}
\toprule
      & \multicolumn{3}{c}{Best Aug.}                             & \multicolumn{3}{c}{Aya-101}                                              & \multicolumn{3}{c}{Multilingual }                             \\     & \multicolumn{3}{c}{Indiv.}                             & \multicolumn{3}{c}{}                                              & \multicolumn{3}{c}{Baseline}                              \\ \cmidrule(lr){2-4} \cmidrule(lr){5-7} \cmidrule(lr){8-10}
Lang. & \multicolumn{1}{l}{R1} & \multicolumn{1}{l}{R2} & \multicolumn{1}{l}{RL} & \multicolumn{1}{l}{R1} & \multicolumn{1}{l}{R2} & \multicolumn{1}{l}{RL} & \multicolumn{1}{l}{R1} & \multicolumn{1}{l}{R2} & \multicolumn{1}{l}{RL} \\

\cmidrule(){1-1} \cmidrule(lr){2-4} \cmidrule(lr){5-7} \cmidrule(lr){8-10}
ckb   & \textbf{52.6}                   & \textbf{38.1}                   & \textbf{44.4}          & 23.9                   & 10.1                   & 18.2                   & 45.2                   & 27.0                   & 34.7                   \\
hat   & 28.7                   & 11.2                   & 20.9                   & 21.5                   & 7.4                    & 16.9                   & \textbf{29.0}                  & \textbf{12.1}                   & \textbf{21.6}          \\
hye   & 34.5                   & 18.3                   & 25.9                   & 27.3                   & 12.2                   & 20.9                   & \textbf{34.9}                   & \textbf{19.2}                   & \textbf{26.5}          \\
kat   & 28.1                   & 13.1                   & 22.9                   & 23.9                   & 9.0                    & 19.8                   & \textbf{29.2}                   & \textbf{14.4}                   & \textbf{24.2}          \\
khm   & 35.1                   & 16.0                   & 28.5                   & 15.6                   & 4.3                    & 13.8                   & \textbf{38.2}                   & \textbf{19.0}                   & \textbf{31.3}          \\
kmr   & 30.5                   & 15.0                   & 23.9                   & 24.6                   & 10.0                   & 19.3                   & \textbf{33.7}                   & \textbf{18.0}                   & \textbf{27.0}          \\
mkd   & 27.0                   & 10.1                   & 20.4                   & 23.9                   & 8.7                    & 18.4                   & \textbf{27.3}                   & \textbf{11.4}                   & \textbf{20.9}          \\\bottomrule
\end{tabular}
}
\caption{Comparison of best augmented individual model with LLM Aya-101 and baseline multilingual model.}
\label{tab:compare-ind}
\end{table}

\paragraph{Multilingual Models}
Although augmented training of individual models performs better than training individual models without augmented data, the scores of individual models are still lower than multilingual fine-tuning of mT5 with a combination of all LR-Sum and XL-Sum training data as seen in Table \ref{tab:compare-ind}. 
However, these individual models do outperform the best performing smaller LLM, Aya-101. 
\citet{hasan-etal-2021-xl} and \citet{palen-michel-lignos-2023-lr} found multilingual models to generally perform better than individually trained models. 
We compare the performance of the best augmented training approach with the reported multilingual model scores from LR-Sum. 

\begin{figure*}[tb]
\centering
\small
    \includegraphics[width=\linewidth]{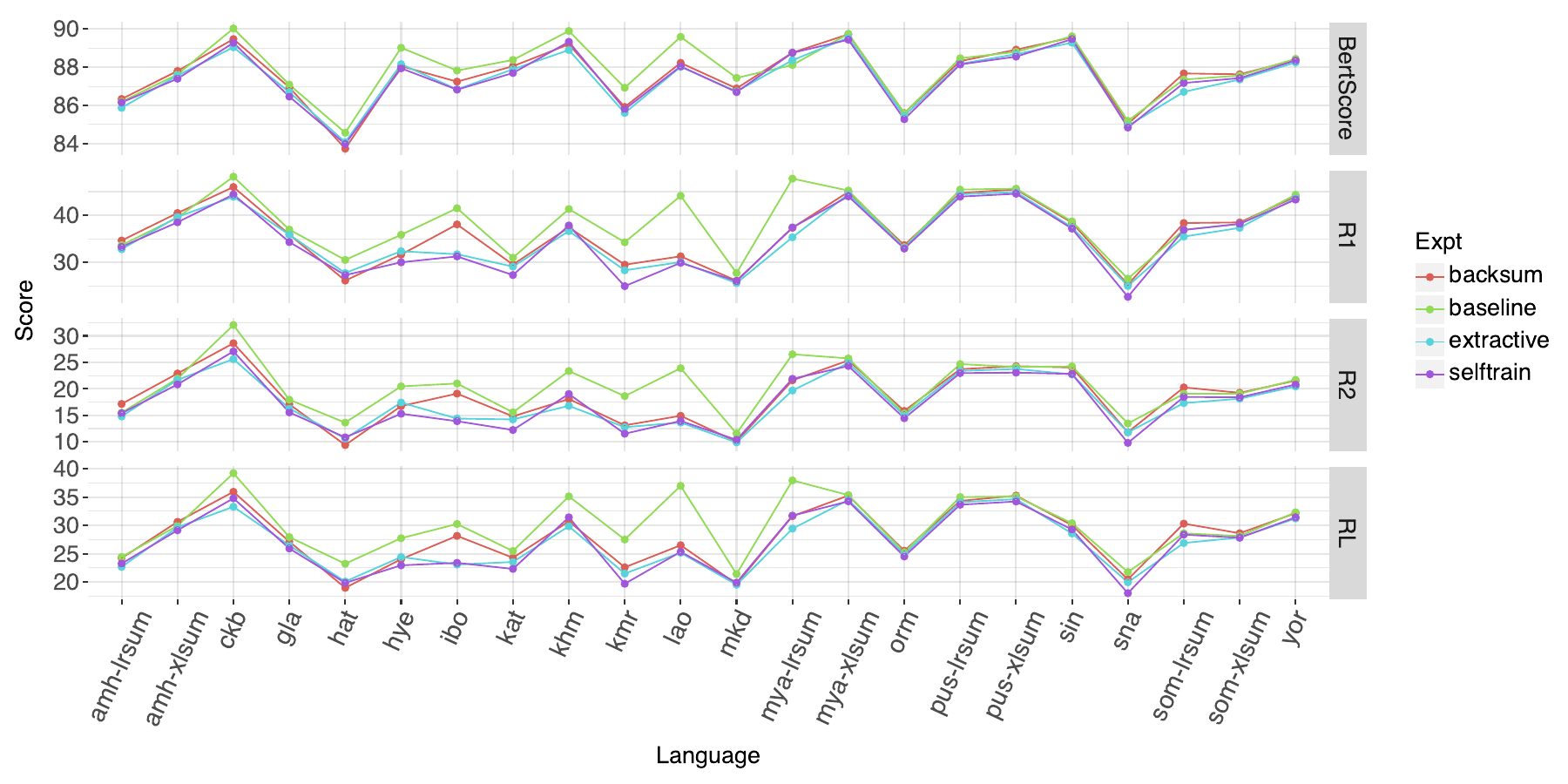}
    \caption{Scores for augmentation approaches with multilingual fine-tuning by language}
    \label{fig:multilingual-scores}
\end{figure*}

The results of including augmented data in the fine-tuning of the multilingual model, shown in Figure \ref{fig:multilingual-scores}, and in more detail in Table \ref{tab:aug-results} in Appendix \ref{app:results}
do not demonstrate a clear improvement over the baseline. 
For Amharic, Sorani, Georgian, Pashto, and Somali, the Back-Summarization approach performs somewhat better. 
Self-training tends to have the same or lower ROUGE Scores for all languages and test set varieties tested. 
Compared to the best performing LLM with prompting, both the multilingual fine-tuning of mT5 with and without augmentation have higher scores.

\paragraph{Smaller LLMs}

\begin{table*}[tb]
\centering 

\resizebox{\textwidth}{!}{%
\begin{tabular}{@{}llrrrrrrrrl@{}}
\toprule
 \multicolumn{2}{c}{} & \multicolumn{3}{c}{Mixtral} & \multicolumn{3}{c}{LLama 3} & \multicolumn{3}{c}{Aya-101} \\ \cmidrule(lr){3-5}\cmidrule(lr){6-8} \cmidrule(lr){9-11}
Dataset & Lang. & \multicolumn{1}{l}{\% Eng.} & \multicolumn{1}{l}{RL} & \multicolumn{1}{l}{BERTScore} & \multicolumn{1}{l}{\% Eng.} & \multicolumn{1}{l}{RL} & \multicolumn{1}{l}{BERTScore} & \multicolumn{1}{l}{\% Eng.} & \multicolumn{1}{l}{RL} & BERTScore \\
\cmidrule(r){1-2}\cmidrule(lr){3-5}\cmidrule(lr){6-8} \cmidrule(lr){9-11}
LR-Sum & amh & 64.6 & 6.96$^{\pm0.02}$ & 80.15$^{\pm0.01}$ & 38.6 & 12.49$^{\pm0.02}$ & 81.83$^{\pm0.01}$ & 0.0 & \textbf{14.63$^{\pm0.03}$} & \textbf{83.29$^{\pm0.01}$} \\
LR-Sum & ckb & 55.8 & 6.28$^{\pm0.01}$ & 78.26$^{\pm0.01}$ & 16.8 & \textbf{29.79$^{\pm0.04}$} & \textbf{86.93$^{\pm0.01}$} & 0.0 & 18.20$^{\pm0.03}$ & 84.56$^{\pm0.01}$ \\
LR-Sum & hat & 2.6 & 13.36$^{\pm0.01}$ & 82.55$^{\pm0.00}$ & 0.2 & \textbf{18.65$^{\pm0.02}$} & \textbf{83.87$^{\pm0.00}$} & 0.6 & 16.86$^{\pm0.02}$ & 82.77$^{\pm0.01}$ \\
LR-Sum & hye & 60.7 & 7.17$^{\pm0.01}$ & 81.16$^{\pm0.02}$ & 4.0 & 14.09$^{\pm0.01}$ & 85.21$^{\pm0.00}$ & 0.0 & \textbf{20.89$^{\pm0.02}$} & \textbf{87.67$^{\pm0.01}$} \\
LR-Sum & kat & 75.4 & 4.83$^{\pm0.01}$ & 81.07$^{\pm0.01}$ & 45.0 & 10.69$^{\pm0.01}$ & 83.18$^{\pm0.01}$ & 0.0 & \textbf{19.85$^{\pm0.02}$} & \textbf{87.84$^{\pm0.01}$} \\
LR-Sum & khm & 76.2 & 4.77$^{\pm0.01}$ & 80.29$^{\pm0.01}$ & 37.9 & 8.77$^{\pm0.01}$ & 81.80$^{\pm0.00}$ & 0.1 & \textbf{13.77$^{\pm0.02}$} & \textbf{86.05$^{\pm0.00}$} \\
LR-Sum & kmr & 5.4 & 12.53$^{\pm0.01}$ & 83.64$^{\pm0.00}$ & 0.0 & 18.76$^{\pm0.02}$ & \textbf{85.28$^{\pm0.00}$} & 0.2 & \textbf{19.25$^{\pm0.02}$} & 84.97$^{\pm0.01}$ \\
LR-Sum & lao & 63.9 & 4.85$^{\pm0.01}$ & 79.61$^{\pm0.01}$ & 49.2 & 10.07$^{\pm0.01}$ & 81.64$^{\pm0.00}$ & 0.2 & \textbf{20.86$^{\pm0.02}$} & \textbf{87.40$^{\pm0.00}$} \\
LR-Sum & mkd & 6.6 & 8.24$^{\pm0.01}$ & 83.52$^{\pm0.01}$ & 0.3 & 14.85$^{\pm0.01}$ & 86.18$^{\pm0.00}$ & 0.0 & \textbf{18.40$^{\pm0.02}$} & \textbf{86.95$^{\pm0.01}$} \\
LR-Sum & mya & 58.6 & 4.09$^{\pm0.00}$ & 79.19$^{\pm0.01}$ & 54.0 & 8.88$^{\pm0.01}$ & 82.00$^{\pm0.00}$ & 0.1 & \textbf{13.79$^{\pm0.01}$} & \textbf{85.67$^{\pm0.00}$} \\
LR-Sum & pus & 47.4 & 9.40$^{\pm0.01}$ & 80.78$^{\pm0.01}$ & 3.2 & 22.57$^{\pm0.01}$ & 85.85$^{\pm0.00}$ & 0.0 & \textbf{28.69$^{\pm0.01}$} & \textbf{87.25$^{\pm0.00}$} \\
LR-Sum & sna & 3.0 & 11.73$^{\pm0.01}$ & 82.13$^{\pm0.00}$ & 0.0 & 15.70$^{\pm0.02}$ & 83.30$^{\pm0.00}$ & 0.2 & \textbf{15.74$^{\pm0.02}$} & \textbf{83.48$^{\pm0.01}$} \\
LR-Sum & som & 13.9 & 12.54$^{\pm0.03}$ & 83.38$^{\pm0.01}$ & 1.8 & \textbf{22.05$^{\pm0.03}$} & \textbf{86.00$^{\pm0.01}$} & 0.0 & 20.98$^{\pm0.04}$ & 85.60$^{\pm0.01}$ \\
\cmidrule(r){1-2}\cmidrule(lr){3-5}\cmidrule(lr){6-8} \cmidrule(lr){9-11}
XL-Sum & amh & 67.4 & 5.62$^{\pm0.01}$ & 79.54$^{\pm0.01}$ & 43.7 & 11.90$^{\pm0.01}$ & 82.01$^{\pm0.00}$ & 0.0 & \textbf{20.88$^{\pm0.02}$} & \textbf{85.21$^{\pm0.01}$} \\
XL-Sum & gla & 2.0 & 13.58$^{\pm0.01}$ & 83.60$^{\pm0.00}$ & 0.2 & 18.54$^{\pm0.01}$ & 84.67$^{\pm0.00}$ & 0.0 & \textbf{26.79$^{\pm0.02}$} & \textbf{86.56$^{\pm0.00}$} \\
XL-Sum & ibo & 13.8 & 13.24$^{\pm0.01}$ & 82.47$^{\pm0.01}$ & 0.0 & 19.55$^{\pm0.01}$ & 85.09$^{\pm0.00}$ & 0.0 & \textbf{29.39$^{\pm0.02}$} & \textbf{87.33$^{\pm0.00}$} \\
XL-Sum & mya & 53.3 & 8.86$^{\pm0.01}$ & 81.45$^{\pm0.01}$ & 38.7 & 14.92$^{\pm0.01}$ & 83.18$^{\pm0.00}$ & 0.0 & \textbf{27.09$^{\pm0.02}$} & \textbf{88.05$^{\pm0.01}$} \\
XL-Sum & orm & 16.9 & 10.38$^{\pm0.01}$ & 81.23$^{\pm0.00}$ & 0.7 & 13.91$^{\pm0.01}$ & 82.47$^{\pm0.01}$ & 1.6 & \textbf{19.48$^{\pm0.01}$} & \textbf{83.28$^{\pm0.00}$} \\
XL-Sum & pus & 40.6 & 10.15$^{\pm0.01}$ & 81.40$^{\pm0.01}$ & 4.0 & 21.37$^{\pm0.01}$ & 85.63$^{\pm0.00}$ & 0.0 & \textbf{30.55$^{\pm0.01}$} & \textbf{87.86$^{\pm0.00}$} \\
XL-Sum & sin & 59.4 & 5.10$^{\pm0.01}$ & 78.82$^{\pm0.03}$ & 15.7 & 14.60$^{\pm0.01}$ & 84.73$^{\pm0.00}$ & 0.0 & \textbf{22.87$^{\pm0.03}$} & \textbf{88.16$^{\pm0.01}$} \\
XL-Sum & som & 17.2 & 11.04$^{\pm0.01}$ & 82.58$^{\pm0.00}$ & 0.8 & 17.61$^{\pm0.01}$ & 84.74$^{\pm0.00}$ & 0.0 & \textbf{22.10$^{\pm0.02}$} & \textbf{85.92$^{\pm0.00}$} \\
XL-Sum & yor & 17.3 & 13.79$^{\pm0.01}$ & 82.85$^{\pm0.00}$ & 0.9 & 19.47$^{\pm0.01}$ & 85.56$^{\pm0.00}$ & 0.1 & \textbf{26.92$^{\pm0.02}$} & \textbf{86.91$^{\pm0.00}$} \\ \bottomrule
\end{tabular}
}
\caption{LLM performance across languages measured in Rouge-L and BERTScore along with the percentage of generated summary containing English found by language id. All results are a a mean of 500 bootstrap samples with standard error reported.}
\label{tab:llm}

\end{table*}

Despite impressive summarization capabilities of LLMs as discussed in \citet{zhang-etal-2024-benchmarking}, the LLM models we explored here performed underwhelmingly.
As seen in Table~\ref{tab:llm}, Mixtral performed the worst while Llama3 tended to perform somewhat better than Mixtral, and  Aya-101 performed best of the LLMs examined. \footnote{Rouge-1 and Rouge-2 are included in Appendix \ref{app:results} in Table \ref{tab:llm-extra}.}
We observe that Mixtral and Llama3 tend to produce a decent amount of English. 
The proportion of English appears to be highest when the target language has a non-Roman script. 
For example, Amharic, Georgian, Khmer, Lao, and Burmese all produce English with mean proportions over 30\%. 

We manually reviewed the responses of LLMs when they generated English output despite being given non-English articles and being prompted to respond in the target language.
Sometimes the English response is an apology message explaining that it cannot perform the task;
other times it is a plausible English summary of the target article. 
In some cases the model asks to see the text despite having already been shown the text, and sometimes the model begins in the target language but eventually switches to English.  

\section{Additional Experiments}

This work was developed over years %
by a research group with relatively limited computational resources.
During development of this work, we gained access to GPUs with more memory, new LLMs were released, and better LLM as judge approaches were developed.
These factors led us to perform an additional set of experiments that builds upon the results in the previous section but explores using larger LLMs and using LLMs in additional ways, namely for evaluation and for a translate-summarize-translate pipeline.\footnote{We separate this work into a different section, as these experiments were able to be run with more languages and with additional evaluation metrics.}
So now, having compared mT5 fine-tuning methods with smaller LLMs and finding the multilingual baseline mT5 model and Aya-101 with highest ROUGE and BERTScores, we turn to using larger LLMs and evaluating with M-Prometheus.

\subsection{Methods}
\paragraph{Larger LLMs}
 We additionally run inference using 3 moderately sized LLMs, Gemma-3 27B \citep{gemma_2025}, Aya-Expanse 32B \citep{dang2024ayaexpansecombiningresearch}, and  Llama 3.3 70B. 
 For these runs we use VLLM for inference. 
 The prompts we used request that the model generate a summary in two sentences. 
 Two sentences of summary was requested to be similar to the reference summaries for the datasets used given that most of the reference summaries are roughly two sentences long. 
The specific prompts used are detailed in Appendix \ref{sec:prompts}.

\paragraph{Translate-Summarize-Translate (TST) with LLMs}
We experiment with a pipeline approach to generating summaries in less-resourced languages where an LLM is first prompted to translate the article into English, and then summarize in two sentences the English-translated article. 
Finally, the model is prompted to translate back to the target language.
We run experiments using Gemma-3 27B and Aya-Expanse 32B since these two LLMs had better performance than Llama 3.3 when doing simple summarization prompts.
The specific prompts used are detailed in Appendix \ref{sec:prompts}.

\subsection{Results}

  \begin{table*}[tb]
 \centering
 \footnotesize
\begin{tabular}{@{}llrrrrrr@{}}
\toprule
        &      & \multicolumn{2}{c}{Aya-Expanse 32b}    &                    \multicolumn{2}{c}{Gemma-3}                       & \multicolumn{2}{c}{Llama 3.3}   \\                 
        \cmidrule(lr){3-4}\cmidrule(lr){5-6} \cmidrule(lr){7-8} 
Dataset & Lang. & \multicolumn{1}{c}{RL}                & \multicolumn{1}{c}{BERTScore}         & \multicolumn{1}{c}{RL}                & \multicolumn{1}{c}{BERTScore}         & \multicolumn{1}{c}{RL}                & \multicolumn{1}{c}{BERTScore}         \\
\midrule
LR-Sum  & amh  & 20.54$^{\pm0.03}$ & 85.14$^{\pm0.01}$ & 16.88$^{\pm0.03}$ & 84.03$^{\pm0.02}$ & \textbf{21.83}$^{\pm0.03}$ & \textbf{85.68}$^{\pm0.01}$ \\
LR-Sum  & ckb  & \textbf{21.94}$^{\pm0.02}$ & \textbf{85.18}$^{\pm0.00}$ & 15.81$^{\pm0.01}$ & 83.48$^{\pm0.01}$ & 21.26$^{\pm0.02}$ & 84.69$^{\pm0.01}$ \\
LR-Sum  & hat  & \textbf{20.48}$^{\pm0.02}$ & \textbf{84.25}$^{\pm0.00}$ & 17.02$^{\pm0.01}$ & 83.39$^{\pm0.00}$ & 20.09$^{\pm0.02}$ & 84.12$^{\pm0.00}$ \\
LR-Sum  & hye  & 19.76$^{\pm0.01}$ & 86.91$^{\pm0.00}$ & 19.67$^{\pm0.02}$ & 87.70$^{\pm0.00}$ & \textbf{20.48}$^{\pm0.02}$ & \textbf{87.76}$^{\pm0.00}$ \\
LR-Sum  & kat  & 17.75$^{\pm0.02}$ & 86.99$^{\pm0.00}$ & 17.05$^{\pm0.01}$ & 86.16$^{\pm0.02}$ & \textbf{20.12}$^{\pm0.02}$ & \textbf{87.63}$^{\pm0.00}$ \\
LR-Sum  & khm  & 13.03$^{\pm0.01}$ & 85.33$^{\pm0.00}$ & 13.10$^{\pm0.01}$ & 85.50$^{\pm0.01}$ & \textbf{15.02}$^{\pm0.01}$ & \textbf{85.91}$^{\pm0.00}$ \\
LR-Sum  & kmr  & 15.45$^{\pm0.02}$ & 82.80$^{\pm0.01}$ & 15.54$^{\pm0.01}$ & 84.45$^{\pm0.01}$ & \textbf{19.70}$^{\pm0.02}$ & \textbf{85.88}$^{\pm0.00}$ \\
LR-Sum  & lao  & 21.36$^{\pm0.01}$ & 87.06$^{\pm0.00}$ & 12.46$^{\pm0.01}$ & 84.69$^{\pm0.00}$ & \textbf{23.52}$^{\pm0.01}$ & \textbf{87.78}$^{\pm0.00}$ \\
LR-Sum  & mkd  & 16.07$^{\pm0.01}$ & 85.87$^{\pm0.00}$ & 14.94$^{\pm0.01}$ & 86.24$^{\pm0.01}$ & \textbf{16.35}$^{\pm0.01}$ & \textbf{86.70}$^{\pm0.00}$ \\
LR-Sum  & mya  & \textbf{23.88}$^{\pm0.02}$ & \textbf{87.49}$^{\pm0.00}$ & 11.63$^{\pm0.01}$ & 82.75$^{\pm0.02}$ & 18.08$^{\pm0.02}$ & 87.16$^{\pm0.00}$ \\
LR-Sum  & pus  & 24.82$^{\pm0.01}$ & 86.13$^{\pm0.00}$ & 21.05$^{\pm0.01}$ & 85.67$^{\pm0.00}$ & \textbf{25.03}$^{\pm0.01}$ & \textbf{87.11}$^{\pm0.00}$ \\
LR-Sum  & sna  & 16.58$^{\pm0.02}$ & 83.89$^{\pm0.00}$ & 14.73$^{\pm0.01}$ & 83.26$^{\pm0.00}$ & \textbf{16.72}$^{\pm0.02}$ & \textbf{83.48}$^{\pm0.00}$ \\
LR-Sum  & som  & \textbf{23.83}$^{\pm0.03}$ & 86.71$^{\pm0.01}$ & 19.62$^{\pm0.03}$ & 85.85$^{\pm0.01}$ & 23.35$^{\pm0.03}$ & \textbf{86.88}$^{\pm0.01}$ \\
\bottomrule
\end{tabular}
\caption{Performance of Larger LLMs with simple prompting for summaries in 2 sentences}
\label{tab:larger-llms}
\end{table*}

\begin{figure*}[tb!]
\centering
\small
    \includegraphics[width=\linewidth]{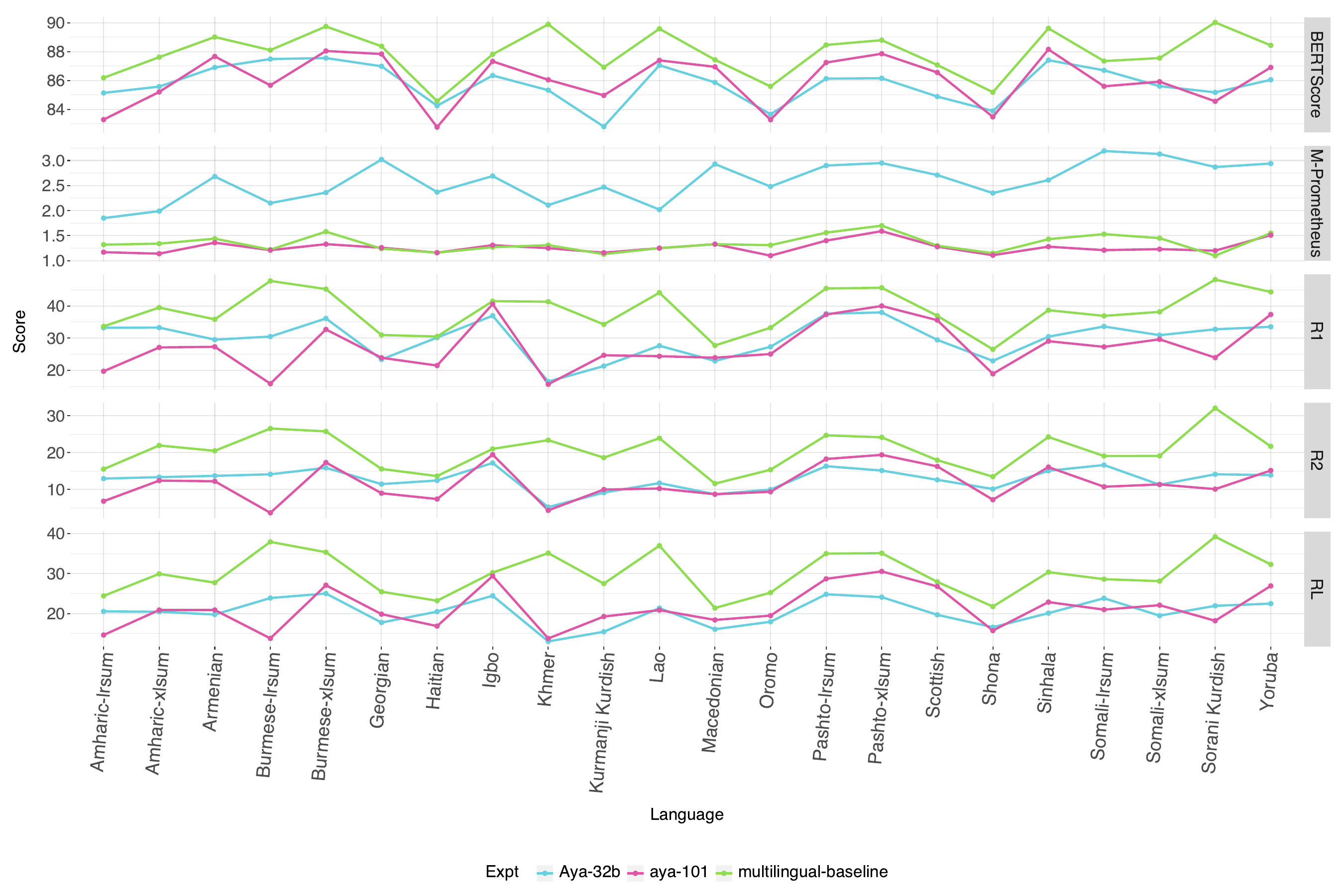}
    \caption{Comparison of Aya-101, Aya-Expanse, and Multilingual Transfer MT5 finetuned baseline}
    \label{fig:ayas-vs-multibaseline}
\end{figure*}

 \paragraph{Larger LLMs}
When comparing relatively larger LLMs of similar sizes, shown in Table~\ref{tab:larger-llms},
Aya-Expanse and Llama 3.3 tend to outperform Gemma, but neither completely outperforms the other for all languages. 
All smaller LLMs performed worse than larger LLMs with the exception of Llama 3's performance on Sorani Kurdish and Aya-101, which has better performance in some languages than larger LLMs.
We see in Figure~\ref{fig:ayas-vs-multibaseline} that Aya-Expanse tends to have scores below the multilingual fine-tuned mT5 baseline model, with the exception of M-Prometheus scores which tend to favor Aya-Expanse 32b.
Aya-101 and Aya-Expanse are comparable in most cases, but it varies based on metric and language.

\paragraph{Translate-Summarize-Translate (TST)}

\begin{table}[tbh]
 \centering
\footnotesize
 \resizebox{\columnwidth}{!}{%
\begin{tabular}{@{}lrrrrrrr@{}}
\toprule
         & \multicolumn{2}{c}{Aya Exp. TST}                            & \multicolumn{2}{c}{Gemma TST}                          & \multicolumn{2}{c}{Best Non-TST}                       \\ \cmidrule(lr){2-3}\cmidrule(lr){4-5} \cmidrule(lr){6-7} 
 Lang & \multicolumn{1}{l}{RL} & \multicolumn{1}{l}{BS} & \multicolumn{1}{l}{RL} & \multicolumn{1}{l}{BS} & \multicolumn{1}{l}{RL} & \multicolumn{1}{l}{BS} \\
 \cmidrule(r){1-1}\cmidrule(lr){2-3}\cmidrule(lr){4-5} \cmidrule(lr){6-7} 
   amh  & 10.79                  & 80.86                         & 15.29                  & 83.98                         & \textbf{21.83}         & \textbf{85.68}                \\
   ckb  & 13.62                  & 82.95                         & 12.44                  & 82.18                         & \textbf{21.94}         & \textbf{85.18}                \\
   hat  & 12.98                  & 82.62                         & 14.57                  & 83.15                         & \textbf{20.48}         & \textbf{84.25}                \\
 hye  & 11.68                  & 83.64                         & 14.84                  & 86.49                         & \textbf{20.48}         & \textbf{87.76}                \\
 kat  & 11.86                  & 84.88                         & 12.61                  & 85.96                         & \textbf{20.12}         & \textbf{87.63}                \\
 khm  & 3.49                   & 79.13                         & 10.83                  & 84.96                         & \textbf{15.02}         & \textbf{85.91}                \\
 kmr  & 9.77                   & 82.07                         & 11.69                  & 84.18                         & \textbf{19.70}         & \textbf{85.88}                \\
 lao  & 6.41                   & 81.26                         & 13.25                  & 85.81                         & \textbf{23.52}         & \textbf{87.78}                \\
 mkd  & 12.00                  & 84.27                         & 13.15                  & 86.00                         & \textbf{16.35}         & \textbf{86.70}                \\
 mya  & 8.10                   & 82.15                         & 8.80                   & 84.88                         & \textbf{23.88}         & \textbf{87.49}                \\
 pus  & 14.83                  & 82.33                         & 17.65                  & 85.37                         & \textbf{25.03}         & \textbf{87.11}                \\
 sna  & 8.46                   & 81.66                         & 9.66                   & 81.98                         & \textbf{16.72}         & \textbf{83.48}                \\
 som  & 13.44                  & 83.47                         & 15.61                  & 85.19                         & \textbf{23.83}         & \textbf{86.71}                \\ \bottomrule
\end{tabular}
}
\caption{Results for Translate-Summarize-Translate Pipeline approach with ROUGE-L (RL) and BERTScore (BS).}
\label{tab:tst-results}
\end{table}

Overall, the TST pipeline results (Table~\ref{tab:tst-results}) demonstrated worse performance across all metrics than the highest performing simple zero-shot prompting.\footnote{Standard error included in Appendix~\ref{app:results} Table~\ref{tab:tst-results-st-err}.}
By examining the output, we observed that although smaller LLMs often produced output which was entirely an English response, larger LLMs sometimes produced additional English commentary in addition to a summary in the target language. 
We postprocessed the LLM output to include only the target summary by filtering out extra English commentary using a combination of pattern matching and language ID. 
This English commentary is broken down into categories 
in Appendix~\ref{app:results} Figure~\ref{fig:extra-eng}. 
Gemma produced the most additional commentary, nearly always mentioning a translation when using the TST approach and sometimes providing a translation in the simple summarization prompt approach. 
Languages like Pashto, Sorani Kurdish, and Amharic show higher rates of the LLM providing unprompted transliteration and pronunciation.  

\paragraph{LLM as Judge}
We use M-Prometheus \citep{pombal2025mprometheussuiteopenmultilingual} for LLM as judge evaluation of summaries. 
M-Prometheus is a multilingual version of Prometheus \citep{kim-etal-2024-prometheus}, which is an open-sourced LLM specifically tuned for text generation evaluation. 
M-Prometheus is based on Qwen 2.5 \citep{qwen2} which claims support for roughly 30 languages. 
M-Prometheus is tested on roughly 30 languages, most of which are higher resourced.
The criteria and prompts provided for evaluation are described in the Appendix Section \ref{sec:llm-judge-prompt}.

We find M-Prometheus scores favor larger LLM output such as that of Aya-Expanse 32b (as seen in Figure \ref{fig:ayas-vs-multibaseline}). 
We examined the distribution of summaries' ROUGE-L scores from Llama 3.3, comparing M-Prometheus with BERTScore and ROUGE-1.
We show the distribution of summaries' ROUGE-L scores from Llama 3.3 compared with M-Prometheus scores in Figure \ref{fig:m-prometheus-RL}, and BERT Score and ROUGE 1 in Appendix \ref{sec:eval-figures}, Figures \ref{fig:m-prometheus-R1} and \ref{fig:m-prometheus-bertscore}. 

We observed that M-Prometheus scores generally do not appear to increase with other scoring metrics and that there is a larger variance for less-resourced languages, while M-Prometheus scores on English summaries appear to have a better alignment with reference-based metrics and less variance.
This suggests that M-Prometheus may be more reliable at evaluating on higher-resource than less-resourced languages.
There is some slight pattern of increase between M-Prometheus scores 1-3, but the distribution of other scoring metrics appears to be slightly lower generally for M-Prometheus scores of 4 and 5.

\section{Discussion}

\textbf{Which approaches work best overall for summarization in less-resourced languages?}
The TST pipeline approach did not outperform zero-shot LLM prompting. 
We compare the best zero-shot LLM of smaller models, larger models with the best performing fine-tuning of mT5 in Figure \ref{fig:ayas-vs-multibaseline}.
Multilingual T5 without augmentation out-performed LLMs for all metrics except M-Prometheus.

\textbf{How well do different metrics measure summarization performance?}
While our aim is not to conduct a comprehensive assessment of summarization metrics, we do observe some differences in metrics.
BERTScore, RL, and R1 can show some nuances in performance when comparing languages, but general trends appear to be mostly consistent across these metrics.
Meanwhile, reference free LLM eval with M-Prometheus tends to favor larger LLM model output. 
It is possible that M-Prometheus is better equipped to evaluate nuances in summarization that reference-based summarization miss.
However, it appears M-Prometheus, being largely trained on six languages and evaluated on a set of roughly 30 higher-resourced languages, has a bias towards larger LLMs trained similarly on mostly higher-resourced languages and we found evidence it may be less reliable in measuring less-resourced languages.

\textbf{Which LLMs output the most English and what kind of English output is it?}
Smaller LLMs tend to have English responses that refuse to complete the task or summarize in English. %
Of the smaller models, Aya-101 avoids English responses best. 
We find that larger, more recent LLMs avoid this trend and instead add extra English text in their response. 
Gemma-3 tends to produce extra English comments the most of models examined and the type of comments varies some by language.

\textbf{Single language or multilingual-fine-tuning with synthetic data?}
We find generally, that multilingual fine-tuning still works best with most languages we examined even when synthetic data is added. 
However there is evidence from Sorani Kurdish that scores that the best performing augmentation approach with individual language fine-tuning can outperform the multilingual fine-tuning baseline by a significant margin. 

\textbf{What augmentation approach works best?}
 For individual models, we found that each data augmentation approach showed an increase in ROUGE scores over the baseline, but there was not one approach that proved to be definitely better than any other across languages.
 For multilingual models, back-summarization appeared to be the most competitive augmentation strategy, but the baseline without augmentation performed better for many languages.

\section{Conclusion}
We have compared a variety of different approaches to summarization with less-resourced languages. 
We found that one of the biggest challenges with LLMs is not outputting English, and found that LLMs, and other strategies including data augmentation and a translation pipeline still under-perform a fine-tuned mT5 multilingual baseline with more traditional reference-based metric. LLM judge M-Prometheus shows a preference for LLM generated summaries, and appears less reliable when evaluating less-resourced summaries.

\section*{Limitations}

Some LLMs in our experiments were loaded with quantization due to resource constraints, so is possible could have had higher performance if the non-quantized model could have been used.
\citet{marchisio2024does} demonstrated that quantization can have more prominent impacts on human evaluation than automatic metrics and that not all languages are impacted equally with multilingual models.
Unfortunately, we were resource constrained and could only make of the LLMs in a quantized setting.

An important limitation to this work is that the evaluation is done entirely with automated metrics.
Limitations to summarization metrics are known and human evaluation is preferred. 
However, human evaluation can be expensive and especially difficult for less-resourced languages due to the added difficulty in recruitment of annotators and quality control with a team of speakers of a diverse set of languages.
We have done our best to report reasonable evaluation metrics and release our model generated summaries for further evaluation in future work by speakers of these languages.
We unfortunately did not have funding for the thousands of dollars required to perform a substantial human evaluation even for a subset of the languages we study in this work.

Our work is limited to claims about the particular models and datasets studied. 
While we examined multiple languages, data augmentation strategies, and LLMs, it is possible that the findings we observed here may not be the same as those on a different set of languages or different LLMs.
However, we believe that our experiments and observations are still informative and of interest to the research community.

\section*{Ethical Considerations and Broader Impact}
Like any text generation model, automatic summarization is based on statistical properties of language and is likely to sometimes generate statements that may be false. 
The models and approaches described in this work are primarily for research purposes and summaries generated by these models are only intended to be used to aid human creation of summaries and should be viewed with skepticism regarding their factual content. 

The datasets used in this work are free and openly available to the public.
While we did not collaborate directly with speakers of the languages studied in this work, we make our model outputs publicly available and welcome collaborations with speakers of the languages studied in order to further investigate approaches to summarization in these languages.

\section*{Bibliographical References}\label{sec:reference}

\bibliographystyle{lrec2026-natbib}
\bibliography{main,anthology}

\section*{Language Resource References}
\label{lr:ref}
\bibliographystylelanguageresource{lrec2026-natbib}
\bibliographylanguageresource{languageresource,anthology}

\input{appendix}

\end{document}

%% file: appendix.tex
\appendix

\section{Results Tables}
 \label{app:results}
 We report ROUGE-1, ROUGE-2, ROUGE-L and BERTScore for experiments with individual models in Table \ref{tab:individual-results} and multilingual models in Table \ref{tab:aug-results}. 
 All results are mean results and include standard error which was computed using bootstrap sampling with 500 resamples \citep{tibshirani1993introduction}. 

\begin{table*}[ht!]
\small
\centering
\resizebox{\textwidth}{!}{%
\begin{tabular}{@{}llllllllllllllllll@{}}
\toprule
 \multicolumn{2}{c}{} & \multicolumn{4}{c}{Baseline} & \multicolumn{4}{c}{Extractive} & \multicolumn{4}{c}{Self-train} & \multicolumn{4}{c}{Backsum} \\ \cmidrule(lr){1-2}\cmidrule(lr){3-6} \cmidrule(lr){7-10} \cmidrule(lr){11-14} \cmidrule(lr){15-18}  
Dataset & Lang. & R1 & R2 & RL & BERTScore & R1 & R2 & RL & BERTScore & R1 & R2 & RL & BERTScore & R1 & R2 & RL & BERTScore \\
\cmidrule(lr){1-2}\cmidrule(lr){3-6} \cmidrule(lr){7-10} \cmidrule(lr){11-14} \cmidrule(lr){15-18}  
lr-sum & ckb & 40.33$^{\pm0.04}$ & 22.00$^{\pm0.05}$ & 29.65$^{\pm0.05}$ & 88.34$^{\pm0.01}$ & \textbf{51.78$^{\pm0.05}$} & \textbf{37.69$^{\pm0.06}$} & \textbf{43.52$^{\pm0.06}$} & \textbf{90.54$^{\pm0.01}$} & 46.45$^{\pm0.05}$ & 30.04$^{\pm0.06}$ & 37.18$^{\pm0.05}$ & 89.64$^{\pm0.01}$ & 47.49$^{\pm0.04}$ & 31.18$^{\pm0.06}$ & 37.95$^{\pm0.05}$ & 89.75$^{\pm0.01}$ \\
lr-sum & hat & 24.79$^{\pm0.03}$ & 9.93$^{\pm0.02}$ & 18.65$^{\pm0.02}$ & 82.53$^{\pm0.00}$ & 28.18$^{\pm0.03}$ & 11.25$^{\pm0.03}$ & 20.54$^{\pm0.03}$ & 84.11$^{\pm0.01}$ & 27.96$^{\pm0.03}$ & \textbf{11.53$^{\pm0.03}$} & \textbf{20.80$^{\pm0.03}$} & \textbf{84.14$^{\pm0.00}$} & \textbf{28.20$^{\pm0.03}$} & 11.20$^{\pm0.03}$ & 20.33$^{\pm0.03}$ & \textbf{84.14$^{\pm0.00}$} \\
lr-sum & hye & 28.03$^{\pm0.02}$ & 12.41$^{\pm0.02}$ & 20.20$^{\pm0.02}$ & 87.22$^{\pm0.00}$ & \textbf{33.84$^{\pm0.03}$} & 18.39$^{\pm0.03}$ & \textbf{25.85$^{\pm0.03}$} & 88.30$^{\pm0.01}$ & 33.09$^{\pm0.03}$ & \textbf{18.41$^{\pm0.03}$} & 25.76$^{\pm0.03}$ & \textbf{88.45$^{\pm0.01}$} & 20.16$^{\pm0.01}$ & 5.76$^{\pm0.01}$ & 15.55$^{\pm0.01}$ & 84.55$^{\pm0.00}$ \\
lr-sum & kat & 22.68$^{\pm0.02}$ & 7.99$^{\pm0.02}$ & 18.08$^{\pm0.02}$ & 84.63$^{\pm0.01}$ & 26.58$^{\pm0.03}$ & 11.46$^{\pm0.03}$ & 21.07$^{\pm0.03}$ & 87.45$^{\pm0.01}$ & 27.28$^{\pm0.04}$ & 12.20$^{\pm0.04}$ & 21.99$^{\pm0.04}$ & 87.77$^{\pm0.01}$ & \textbf{28.09$^{\pm0.04}$} & \textbf{13.13$^{\pm0.04}$} & \textbf{22.94$^{\pm0.04}$} & \textbf{87.85$^{\pm0.01}$} \\
lr-sum & khm & 34.21$^{\pm0.04}$ & 15.15$^{\pm0.04}$ & 27.49$^{\pm0.04}$ & 88.42$^{\pm0.01}$ & 34.62$^{\pm0.03}$ & 14.87$^{\pm0.04}$ & 27.41$^{\pm0.03}$ & 88.44$^{\pm0.01}$ & \textbf{34.95$^{\pm0.04}$} & 16.03$^{\pm0.04}$ & \textbf{28.45$^{\pm0.04}$} & \textbf{88.59$^{\pm0.01}$} & 34.84$^{\pm0.04}$ & \textbf{16.16$^{\pm0.04}$} & 28.40$^{\pm0.04}$ & 88.52$^{\pm0.01}$ \\
lr-sum & kmr & 25.59$^{\pm0.02}$ & 9.41$^{\pm0.02}$ & 20.17$^{\pm0.02}$ & 83.96$^{\pm0.00}$ & \textbf{30.56$^{\pm0.03}$} & 14.40$^{\pm0.04}$ & 23.53$^{\pm0.03}$ & 85.92$^{\pm0.01}$ & 30.40$^{\pm0.04}$ & \textbf{14.94$^{\pm0.04}$} & \textbf{23.80$^{\pm0.04}$} & \textbf{86.17$^{\pm0.01}$} & 29.07$^{\pm0.03}$ & 13.22$^{\pm0.04}$ & 22.28$^{\pm0.04}$ & 85.76$^{\pm0.01}$ \\
lr-sum & mkd & 26.04$^{\pm0.03}$ & 10.28$^{\pm0.03}$ & 19.47$^{\pm0.03}$ & 86.56$^{\pm0.00}$ & \textbf{26.52$^{\pm0.03}$} & 10.21$^{\pm0.03}$ & 19.79$^{\pm0.03}$ & \textbf{86.69$^{\pm0.00}$} & 26.06$^{\pm0.03}$ & 10.24$^{\pm0.03}$ & 19.74$^{\pm0.03}$ & 86.65$^{\pm0.00}$ & 26.41$^{\pm0.03}$ & \textbf{10.84$^{\pm0.03}$} & \textbf{19.85$^{\pm0.03}$} & 86.67$^{\pm0.00}$ \\ \bottomrule
\end{tabular}
}
\caption{Results of data augmentation experiments for individual models for each language. Results are a single run.}
\label{tab:individual-results}
\end{table*}
 
\begin{table*}[h!]
\centering
\small
\resizebox{\textwidth}{!}{%
\begin{tabular}{@{}llllllllllllllllll@{}}
\toprule
 &  & \multicolumn{4}{l}{Baseline} & \multicolumn{4}{l}{Extractive} & \multicolumn{4}{l}{Self-Train} & \multicolumn{4}{l}{Backsum} \\ 
 \cmidrule(lr){1-2}\cmidrule(lr){3-6} \cmidrule(lr){7-10} \cmidrule(lr){11-14} \cmidrule(lr){15-18}  
\textbf{Dataset} & \textbf{Lang.} & \textbf{R1} & \textbf{R2} & \textbf{RL} & \textbf{BERTScore} & \textbf{R1} & \textbf{R2} & \textbf{RL} & \textbf{BERTScore} & \textbf{R1} & \textbf{R2} & \textbf{RL} & \textbf{BERTScore} & \textbf{R1} & \textbf{R2} & \textbf{RL} & \textbf{BERTScore} \\
\cmidrule(lr){1-2}\cmidrule(lr){3-6} \cmidrule(lr){7-10} \cmidrule(lr){11-14} \cmidrule(lr){15-18}  
lr-sum & amh & 33.64$^{\pm0.05}$ & 15.50$^{\pm0.06}$ & 24.42$^{\pm0.05}$ & 86.19$^{\pm0.01}$ & 32.76$^{\pm0.04}$ & 14.78$^{\pm0.05}$ & 22.65$^{\pm0.04}$ & 85.87$^{\pm0.01}$ & 33.27$^{\pm0.05}$ & 15.39$^{\pm0.05}$ & 23.24$^{\pm0.04}$ & 86.15$^{\pm0.01}$ & \textbf{34.62$^{\pm0.05}$} & \textbf{17.14$^{\pm0.05}$} & \textbf{24.21$^{\pm0.05}$} & \textbf{86.34$^{\pm0.01}$} \\
lr-sum & ckb & 48.18$^{\pm0.05}$ & 32.04$^{\pm0.06}$ & 39.22$^{\pm0.06}$ & \textbf{90.03$^{\pm0.01}$} & 43.92$^{\pm0.04}$ & 25.60$^{\pm0.06}$ & 33.28$^{\pm0.05}$ & 89.04$^{\pm0.01}$ & 44.40$^{\pm0.05}$ & 27.06$^{\pm0.06}$ & 34.78$^{\pm0.05}$ & 89.28$^{\pm0.01}$ & \textbf{45.98$^{\pm0.04}$} & \textbf{28.60$^{\pm0.06}$} & \textbf{35.93$^{\pm0.05}$} & 89.47$^{\pm0.01}$ \\
lr-sum & hat & \textbf{30.49$^{\pm0.03}$} & \textbf{13.62$^{\pm0.03}$} & \textbf{23.21$^{\pm0.03}$} & \textbf{84.57$^{\pm0.01}$} & 27.71$^{\pm0.03}$ & 10.62$^{\pm0.03}$ & 20.06$^{\pm0.03}$ & 84.08$^{\pm0.00}$ & 27.27$^{\pm0.03}$ & 10.84$^{\pm0.03}$ & 19.83$^{\pm0.03}$ & 83.98$^{\pm0.01}$ & 26.10$^{\pm0.03}$ & 9.40$^{\pm0.03}$ & 18.96$^{\pm0.02}$ & 83.73$^{\pm0.00}$ \\
lr-sum & hye & \textbf{35.83$^{\pm0.03}$} & \textbf{20.47$^{\pm0.03}$} & \textbf{27.73$^{\pm0.03}$} & \textbf{89.02$^{\pm0.00}$} & 32.34$^{\pm0.03}$ & 17.40$^{\pm0.03}$ & 24.40$^{\pm0.03}$ & 88.16$^{\pm0.01}$ & 29.99$^{\pm0.03}$ & 15.30$^{\pm0.03}$ & 22.93$^{\pm0.03}$ & 87.94$^{\pm0.01}$ & 31.63$^{\pm0.03}$ & 16.75$^{\pm0.03}$ & 24.02$^{\pm0.03}$ & 88.05$^{\pm0.01}$ \\
lr-sum & kat & \textbf{30.95$^{\pm0.04}$} & \textbf{15.56$^{\pm0.04}$} & \textbf{25.46$^{\pm0.04}$} & \textbf{88.38$^{\pm0.01}$} & 29.08$^{\pm0.04}$ & 14.22$^{\pm0.04}$ & 23.53$^{\pm0.04}$ & 87.89$^{\pm0.01}$ & 27.28$^{\pm0.04}$ & 12.23$^{\pm0.04}$ & 22.30$^{\pm0.04}$ & 87.70$^{\pm0.01}$ & 29.51$^{\pm0.03}$ & 14.80$^{\pm0.04}$ & 24.27$^{\pm0.04}$ & 88.05$^{\pm0.01}$ \\
lr-sum & khm & \textbf{41.30$^{\pm0.05}$} & 23.36$^{\pm0.06}$ & 35.12$^{\pm0.05}$ & 89.90$^{\pm0.01}$ & 36.63$^{\pm0.04}$ & 16.79$^{\pm0.05}$ & 29.83$^{\pm0.04}$ & 88.90$^{\pm0.01}$ & 37.81$^{\pm0.05}$ & 19.02$^{\pm0.05}$ & 31.42$^{\pm0.05}$ & 89.33$^{\pm0.01}$ & 37.40$^{\pm0.04}$ & 18.08$^{\pm0.05}$ & 30.71$^{\pm0.04}$ & 89.20$^{\pm0.01}$ \\
lr-sum & kmr & \textbf{34.25$^{\pm0.04}$} & 18.62$^{\pm0.05}$ & 27.49$^{\pm0.04}$ & 86.92$^{\pm0.01}$ & 28.29$^{\pm0.03}$ & 12.77$^{\pm0.03}$ & 21.48$^{\pm0.03}$ & 85.61$^{\pm0.01}$ & 24.92$^{\pm0.04}$ & 11.53$^{\pm0.04}$ & 19.67$^{\pm0.04}$ & 85.80$^{\pm0.01}$ & 29.48$^{\pm0.03}$ & 13.11$^{\pm0.04}$ & 22.58$^{\pm0.04}$ & 85.92$^{\pm0.01}$ \\
lr-sum & lao & \textbf{44.11$^{\pm0.02}$} & 23.89$^{\pm0.03}$ & 36.98$^{\pm0.02}$ & 89.59$^{\pm0.00}$ & 30.03$^{\pm0.02}$ & 13.64$^{\pm0.02}$ & 25.16$^{\pm0.02}$ & 88.02$^{\pm0.00}$ & 29.88$^{\pm0.02}$ & 13.93$^{\pm0.02}$ & 25.32$^{\pm0.02}$ & 88.04$^{\pm0.00}$ & 31.26$^{\pm0.02}$ & 14.91$^{\pm0.02}$ & 26.48$^{\pm0.02}$ & 88.23$^{\pm0.00}$ \\
lr-sum & mkd & \textbf{27.70$^{\pm0.03}$} & 11.57$^{\pm0.03}$ & 21.39$^{\pm0.03}$ & 87.44$^{\pm0.01}$ & 25.58$^{\pm0.03}$ & 9.89$^{\pm0.03}$ & 19.50$^{\pm0.03}$ & 86.73$^{\pm0.01}$ & 25.99$^{\pm0.03}$ & 10.39$^{\pm0.03}$ & 19.87$^{\pm0.03}$ & 86.70$^{\pm0.01}$ & 26.07$^{\pm0.03}$ & 10.02$^{\pm0.03}$ & 19.63$^{\pm0.03}$ & 86.89$^{\pm0.01}$ \\
lr-sum & mya & \textbf{47.77$^{\pm0.02}$} & 26.52$^{\pm0.02}$ & 37.94$^{\pm0.02}$ & 88.11$^{\pm0.00}$ & 35.30$^{\pm0.03}$ & 19.69$^{\pm0.03}$ & 29.42$^{\pm0.03}$ & 88.37$^{\pm0.01}$ & 37.39$^{\pm0.03}$ & 21.89$^{\pm0.04}$ & 31.71$^{\pm0.03}$ & 88.75$^{\pm0.01}$ & 37.32$^{\pm0.03}$ & 21.60$^{\pm0.03}$ & 31.63$^{\pm0.03}$ & 88.76$^{\pm0.01}$ \\
lr-sum & pus & \textbf{45.44$^{\pm0.01}$} & \textbf{24.67$^{\pm0.01}$} & \textbf{35.00$^{\pm0.01}$} & 88.47$^{\pm0.00}$ & 44.46$^{\pm0.01}$ & 23.38$^{\pm0.01}$ & 34.10$^{\pm0.01}$ & 88.17$^{\pm0.00}$ & 43.94$^{\pm0.01}$ & 22.97$^{\pm0.01}$ & 33.65$^{\pm0.01}$ & 88.14$^{\pm0.00}$ & 44.71$^{\pm0.01}$ & 23.70$^{\pm0.01}$ & 34.32$^{\pm0.01}$ & 88.32$^{\pm0.00}$ \\
lr-sum & sna & \textbf{26.50$^{\pm0.03}$} & \textbf{13.46$^{\pm0.03}$} & \textbf{21.75$^{\pm0.03}$} & \textbf{85.19$^{\pm0.01}$} & 24.96$^{\pm0.03}$ & 11.78$^{\pm0.03}$ & 19.94$^{\pm0.03}$ & 84.93$^{\pm0.01}$ & 22.65$^{\pm0.03}$ & 9.80$^{\pm0.03}$ & 18.01$^{\pm0.03}$ & 84.84$^{\pm0.01}$ & 25.25$^{\pm0.03}$ & 11.80$^{\pm0.03}$ & 20.45$^{\pm0.03}$ & 85.03$^{\pm0.01}$ \\
lr-sum & som & 36.90$^{\pm0.05}$ & 19.05$^{\pm0.05}$ & 28.60$^{\pm0.05}$ & 87.35$^{\pm0.01}$ & 35.44$^{\pm0.05}$ & 17.31$^{\pm0.05}$ & 26.88$^{\pm0.05}$ & 86.71$^{\pm0.01}$ & 36.87$^{\pm0.05}$ & 18.47$^{\pm0.05}$ & 28.36$^{\pm0.05}$ & 87.17$^{\pm0.01}$ & \textbf{38.33$^{\pm0.05}$} & \textbf{20.27$^{\pm0.05}$} & \textbf{30.30$^{\pm0.05}$} & \textbf{87.67$^{\pm0.01}$} \\
xlsum & amh & 39.50$^{\pm0.02}$ & 21.93$^{\pm0.02}$ & 29.93$^{\pm0.02}$ & 87.62$^{\pm0.00}$ & 39.64$^{\pm0.02}$ & 21.69$^{\pm0.02}$ & 29.66$^{\pm0.02}$ & 87.57$^{\pm0.00}$ & 38.49$^{\pm0.02}$ & 20.84$^{\pm0.02}$ & 29.13$^{\pm0.02}$ & 87.40$^{\pm0.00}$ & \textbf{40.46$^{\pm0.02}$} & \textbf{22.89$^{\pm0.02}$} & \textbf{30.62$^{\pm0.02}$} & \textbf{87.80$^{\pm0.00}$} \\
xlsum & gla & \textbf{36.94$^{\pm0.02}$} & \textbf{17.94$^{\pm0.02}$} & \textbf{27.93$^{\pm0.02}$} & \textbf{87.08$^{\pm0.00}$} & 35.85$^{\pm0.02}$ & 16.29$^{\pm0.02}$ & 26.45$^{\pm0.02}$ & 86.69$^{\pm0.00}$ & 34.29$^{\pm0.02}$ & 15.56$^{\pm0.02}$ & 25.91$^{\pm0.02}$ & 86.46$^{\pm0.00}$ & 35.89$^{\pm0.02}$ & 17.10$^{\pm0.02}$ & 27.13$^{\pm0.02}$ & 86.95$^{\pm0.00}$ \\
xlsum & ibo & \textbf{41.48$^{\pm0.03}$} & \textbf{21.00$^{\pm0.02}$} & \textbf{30.24$^{\pm0.02}$} & \textbf{87.82$^{\pm0.00}$} & 31.70$^{\pm0.03}$ & 14.39$^{\pm0.02}$ & 23.08$^{\pm0.02}$ & 86.84$^{\pm0.00}$ & 31.24$^{\pm0.03}$ & 13.89$^{\pm0.02}$ & 23.38$^{\pm0.02}$ & 86.82$^{\pm0.00}$ & 38.04$^{\pm0.03}$ & 19.09$^{\pm0.03}$ & 28.14$^{\pm0.02}$ & 87.24$^{\pm0.01}$ \\
xlsum & mya & \textbf{45.23$^{\pm0.02}$} & \textbf{25.74$^{\pm0.02}$} & \textbf{35.35$^{\pm0.02}$} & \textbf{89.75$^{\pm0.00}$} & 44.47$^{\pm0.02}$ & 24.86$^{\pm0.02}$ & 34.48$^{\pm0.02}$ & 89.54$^{\pm0.00}$ & 44.00$^{\pm0.02}$ & 24.31$^{\pm0.02}$ & 34.24$^{\pm0.02}$ & 89.44$^{\pm0.00}$ & 44.95$^{\pm0.02}$ & 25.36$^{\pm0.02}$ & 35.26$^{\pm0.02}$ & 89.72$^{\pm0.00}$ \\
xlsum & orm & 33.23$^{\pm0.02}$ & 15.34$^{\pm0.02}$ & 25.24$^{\pm0.02}$ & 85.59$^{\pm0.00}$ & 33.13$^{\pm0.02}$ & 15.02$^{\pm0.02}$ & 24.91$^{\pm0.02}$ & 85.45$^{\pm0.00}$ & 32.90$^{\pm0.02}$ & 14.46$^{\pm0.02}$ & 24.50$^{\pm0.01}$ & 85.28$^{\pm0.00}$ & \textbf{33.63$^{\pm0.02}$} & \textbf{15.83$^{\pm0.02}$} & \textbf{25.51$^{\pm0.02}$} & \textbf{85.60$^{\pm0.00}$} \\
xlsum & pus & \textbf{45.64$^{\pm0.01}$} & 24.13$^{\pm0.01}$ & 35.10$^{\pm0.01}$ & 88.80$^{\pm0.00}$ & 44.92$^{\pm0.01}$ & 23.75$^{\pm0.01}$ & 34.62$^{\pm0.01}$ & 88.68$^{\pm0.00}$ & 44.56$^{\pm0.01}$ & 23.06$^{\pm0.01}$ & 34.22$^{\pm0.01}$ & 88.55$^{\pm0.00}$ & 45.47$^{\pm0.01}$ & \textbf{24.28$^{\pm0.01}$} & \textbf{35.25$^{\pm0.01}$} & \textbf{88.92$^{\pm0.00}$} \\
xlsum & sin & \textbf{38.67$^{\pm0.03}$} & \textbf{24.23$^{\pm0.03}$} & \textbf{30.37$^{\pm0.03}$} & \textbf{89.62$^{\pm0.00}$} & 37.43$^{\pm0.03}$ & 22.73$^{\pm0.03}$ & 28.57$^{\pm0.02}$ & 89.28$^{\pm0.00}$ & 37.15$^{\pm0.03}$ & 22.81$^{\pm0.03}$ & 29.26$^{\pm0.02}$ & 89.46$^{\pm0.00}$ & 38.40$^{\pm0.02}$ & 24.02$^{\pm0.03}$ & 30.03$^{\pm0.02}$ & 89.56$^{\pm0.00}$ \\
xlsum & som & 38.15$^{\pm0.02}$ & 19.08$^{\pm0.02}$ & 28.11$^{\pm0.02}$ & 87.56$^{\pm0.00}$ & 37.30$^{\pm0.02}$ & 18.13$^{\pm0.02}$ & 27.85$^{\pm0.02}$ & 87.35$^{\pm0.00}$ & 38.15$^{\pm0.02}$ & 18.40$^{\pm0.02}$ & 27.81$^{\pm0.02}$ & 87.43$^{\pm0.00}$ & \textbf{38.46$^{\pm0.02}$} & \textbf{19.25$^{\pm0.02}$} & \textbf{28.60$^{\pm0.02}$} & \textbf{87.63$^{\pm0.00}$} \\
xlsum & yor & \textbf{44.34$^{\pm0.02}$} & \textbf{21.69$^{\pm0.02}$} & \textbf{32.30$^{\pm0.02}$} & \textbf{88.44$^{\pm0.00}$} & 43.66$^{\pm0.02}$ & 20.44$^{\pm0.02}$ & 31.21$^{\pm0.02}$ & 88.25$^{\pm0.00}$ & 43.31$^{\pm0.02}$ & 20.81$^{\pm0.02}$ & 31.41$^{\pm0.02}$ & 88.35$^{\pm0.00}$ & 43.95$^{\pm0.02}$ & 21.54$^{\pm0.02}$ & 32.15$^{\pm0.02}$ & 88.38$^{\pm0.00}$ \\ \bottomrule
\end{tabular}
}
\caption{Results of multilingual models with data augmentation approaches standard error reported using bootstrapping with 500 samples.}
\label{tab:aug-results}
\end{table*}

\begin{table*}[h!]
\centering 

\resizebox{\textwidth}{!}{%
\begin{tabular}{@{}llrrrrrrrrl@{}}
\toprule
 \multicolumn{2}{c}{} & \multicolumn{3}{c}{Mixtral} & \multicolumn{3}{c}{LLama 3} & \multicolumn{3}{c}{Aya-101} \\ \cmidrule(lr){3-5}\cmidrule(lr){6-8} \cmidrule(lr){9-11}
Dataset & Lang. & \multicolumn{1}{l}{\% Eng.} & \multicolumn{1}{l}{R1} & \multicolumn{1}{l}{R2} & \multicolumn{1}{l}{\% Eng.} & \multicolumn{1}{l}{R1} & \multicolumn{1}{l}{R2} & \multicolumn{1}{l}{\% Eng.} & \multicolumn{1}{l}{R1} & R2 \\
\cmidrule(r){1-2}\cmidrule(lr){3-5}\cmidrule(lr){6-8} \cmidrule(lr){9-11}

LR-Sum & amh & 64.6 & 10.24$^{\pm0.04}$ & 2.59$^{\pm0.02}$ & 38.6 & \textbf{20.04$^{\pm0.04}$} & 5.49$^{\pm0.03}$ & 0.0 & 19.69$^{\pm0.04}$ & \textbf{6.80$^{\pm0.03}$} \\
LR-Sum & ckb & 55.8 & 8.34$^{\pm0.02}$ & 2.37$^{\pm0.01}$ & 16.8 & \textbf{39.40$^{\pm0.04}$} & \textbf{24.87$^{\pm0.05}$} & 0.0 & 23.92$^{\pm0.03}$ & 10.08$^{\pm0.03}$ \\
LR-Sum & hat & 2.6 & 19.61$^{\pm0.02}$ & 9.15$^{\pm0.01}$ & 0.2 & \textbf{26.90$^{\pm0.02}$} & \textbf{12.69$^{\pm0.02}$} & 0.6 & 21.47$^{\pm0.03}$ & 7.39$^{\pm0.02}$ \\
LR-Sum & hye & 60.7 & 10.02$^{\pm0.02}$ & 3.45$^{\pm0.01}$ & 4.0 & 21.85$^{\pm0.02}$ & 7.96$^{\pm0.01}$ & 0.0 & 27.29$^{\pm0.03}$ & \textbf{12.21$^{\pm0.02}$} \\
LR-Sum & kat & 75.4 & 5.83$^{\pm0.01}$ & 1.54$^{\pm0.01}$ & 45.0 & 13.18$^{\pm0.02}$ & 3.67$^{\pm0.01}$ & 0.0 & \textbf{23.88$^{\pm0.03}$} & \textbf{8.97$^{\pm0.03}$} \\
LR-Sum & khm & 76.2 & 5.23$^{\pm0.01}$ & 0.34$^{\pm0.00}$ & 37.9 & 11.57$^{\pm0.01}$ & 1.57$^{\pm0.01}$ & 0.1 & \textbf{15.65$^{\pm0.02}$} & \textbf{4.33$^{\pm0.01}$} \\
LR-Sum & kmr & 5.4 & 16.29$^{\pm0.02}$ & 8.87$^{\pm0.01}$ & 0.0 & \textbf{26.24$^{\pm0.02}$} & \textbf{12.30$^{\pm0.02}$} & 0.2 & 24.65$^{\pm0.03}$ & 9.96$^{\pm0.02}$ \\
LR-Sum & lao & 63.9 & 5.54$^{\pm0.01}$ & 0.80$^{\pm0.00}$ & 49.2 & 12.46$^{\pm0.01}$ & 3.11$^{\pm0.01}$ & 0.2 & \textbf{24.36$^{\pm0.02}$} & \textbf{10.24$^{\pm0.01}$} \\
LR-Sum & mkd & 6.6 & 11.38$^{\pm0.01}$ & 4.54$^{\pm0.01}$ & 0.3 & 20.99$^{\pm0.02}$ & \textbf{8.98$^{\pm0.01}$} & 0.0 & \textbf{23.91$^{\pm0.02}$} & 8.69$^{\pm0.02}$ \\
LR-Sum & mya & 58.6 & 4.65$^{\pm0.01}$ & 0.58$^{\pm0.00}$ & 54.0 & 10.77$^{\pm0.01}$ & 1.72$^{\pm0.00}$ & 0.1 & \textbf{15.83$^{\pm0.01}$} & \textbf{3.65$^{\pm0.01}$} \\
LR-Sum & pus & 47.4 & 12.63$^{\pm0.01}$ & 4.73$^{\pm0.01}$ & 3.2 & 34.32$^{\pm0.01}$ & 15.99$^{\pm0.01}$ & 0.0 & \textbf{37.32$^{\pm0.01}$} & \textbf{18.25$^{\pm0.01}$} \\
LR-Sum & sna & 3.0 & 16.29$^{\pm0.02}$ & 8.48$^{\pm0.01}$ & 0.0 & \textbf{21.06$^{\pm0.02}$} & \textbf{10.68$^{\pm0.02}$} & 0.2 & 18.89$^{\pm0.02}$ & 7.21$^{\pm0.02}$ \\
LR-Sum & som & 13.9 & 17.32$^{\pm0.04}$ & 8.89$^{\pm0.03}$ & 1.8 & \textbf{30.91$^{\pm0.04}$} & \textbf{15.61$^{\pm0.03}$} & 0.0 & 27.27$^{\pm0.04}$ & 10.73$^{\pm0.04}$ \\
\midrule
XL-Sum & amh & 67.4 & 7.38$^{\pm0.01}$ & 1.72$^{\pm0.01}$ & 43.7 & 18.95$^{\pm0.01}$ & 5.30$^{\pm0.01}$ & 0.0 & \textbf{27.09$^{\pm0.02}$} & \textbf{12.39$^{\pm0.02}$} \\
XL-Sum & gla & 2.0 & 20.15$^{\pm0.01}$ & 9.63$^{\pm0.01}$ & 0.2 & 27.29$^{\pm0.02}$ & 11.96$^{\pm0.01}$ & 0.0 & \textbf{35.56$^{\pm0.02}$} & \textbf{16.24$^{\pm0.02}$} \\
XL-Sum & ibo & 13.8 & 18.60$^{\pm0.02}$ & 8.31$^{\pm0.01}$ & 0.0 & 28.64$^{\pm0.02}$ & 14.00$^{\pm0.01}$ & 0.0 & \textbf{40.58$^{\pm0.02}$} & \textbf{19.44$^{\pm0.02}$} \\
XL-Sum & mya & 53.3 & 11.36$^{\pm0.02}$ & 3.33$^{\pm0.01}$ & 38.7 & 20.22$^{\pm0.02}$ & 5.86$^{\pm0.01}$ & 0.0 & \textbf{32.69$^{\pm0.03}$} & \textbf{17.30$^{\pm0.02}$} \\
XL-Sum & orm & 16.9 & 14.99$^{\pm0.01}$ & 5.83$^{\pm0.01}$ & 0.7 & 21.04$^{\pm0.02}$ & 8.28$^{\pm0.01}$ & 1.6 & \textbf{25.04$^{\pm0.02}$} & \textbf{9.37$^{\pm0.01}$} \\
XL-Sum & pus & 40.6 & 14.21$^{\pm0.01}$ & 5.23$^{\pm0.01}$ & 4.0 & 34.14$^{\pm0.01}$ & 14.31$^{\pm0.01}$ & 0.0 & \textbf{39.99$^{\pm0.01}$} & \textbf{19.38$^{\pm0.01}$} \\
XL-Sum & sin & 59.4 & 6.85$^{\pm0.02}$ & 2.00$^{\pm0.01}$ & 15.7 & 22.09$^{\pm0.02}$ & 9.05$^{\pm0.01}$ & 0.0 & \textbf{29.04$^{\pm0.03}$} & \textbf{16.06$^{\pm0.03}$} \\
XL-Sum & som & 17.2 & 16.89$^{\pm0.01}$ & 6.47$^{\pm0.01}$ & 0.8 & 27.75$^{\pm0.01}$ & 9.58$^{\pm0.01}$ & 0.0 & \textbf{29.61$^{\pm0.02}$} & \textbf{11.35$^{\pm0.02}$} \\
XL-Sum & yor & 17.3 & 20.08$^{\pm0.01}$ & 8.35$^{\pm0.01}$ & 0.9 & 29.67$^{\pm0.02}$ & 12.88$^{\pm0.01}$ & 0.1 & \textbf{37.33$^{\pm0.02}$} & \textbf{15.13$^{\pm0.02}$} \\ \bottomrule

\end{tabular}
}
\caption{LLM performance across languages measured in ROUGE-1 and ROUGE-2 along with the percentage of generated summary containing English found by language id. All results are a a mean of 500 bootstrap samples with standard error reported. Rouge-L and BERTScore are included in the main paper, but ROUGE-1 and ROUGE-2 were omitted due to space constraints.}
\label{tab:llm-extra}
\end{table*}

\begin{figure*}[h!]
\centering
\small
    \includegraphics[width=\linewidth]{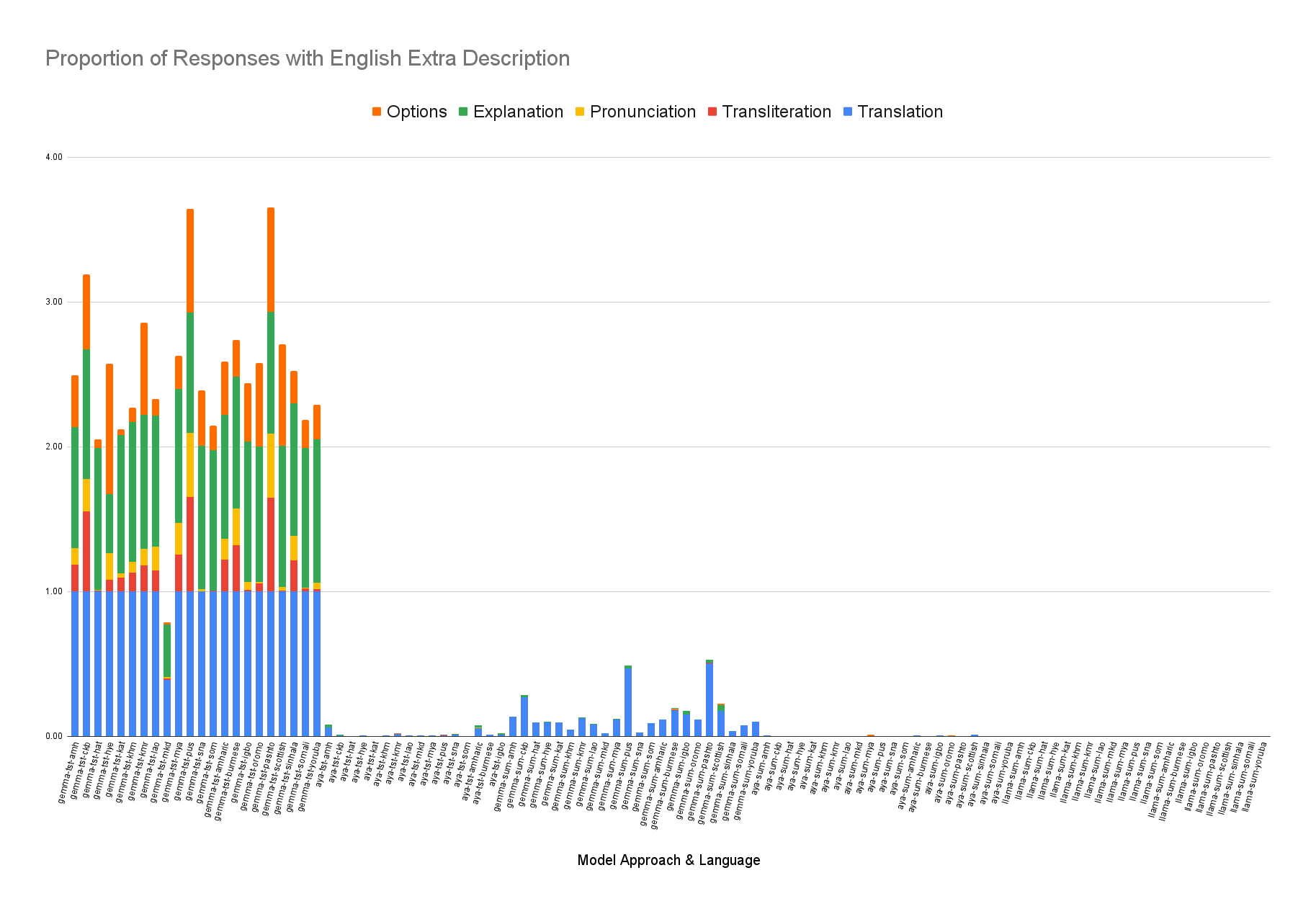}
    \caption{Proportion of summaries with extra English text generated for each model and approach by language by category of Larger LLMs and Translate-Summarize-Translate approach.}
    \label{fig:extra-eng}
\end{figure*}

\begin{table*}[h!]
 \centering
\footnotesize
\begin{tabular}{@{}lllllll@{}}
\toprule
     & \multicolumn{2}{c}{Aya TST}          &                   \multicolumn{2}{c}{Gemma TST}         &                   \multicolumn{2}{c}{Best Non-TST LLM}                                      \\ 
        \cmidrule(lr){2-3}\cmidrule(lr){4-5} \cmidrule(lr){6-7} 
  Lang & RL                & BERTScore         & RL                & BERTScore         & RL                         & BERTScore                  \\
\midrule
 amh  & 10.79$^{\pm0.01}$ & 80.86$^{\pm0.00}$ & 15.29$^{\pm0.02}$ & 83.98$^{\pm0.01}$ & \textbf{21.83$^{\pm0.03}$} & \textbf{85.68$^{\pm0.01}$} \\
 ckb  & 13.62$^{\pm0.01}$ & 82.95$^{\pm0.00}$ & 12.44$^{\pm0.01}$ & 82.18$^{\pm0.00}$ & \textbf{21.94$^{\pm0.02}$} & \textbf{85.18$^{\pm0.00}$} \\
 hat  & 12.98$^{\pm0.01}$ & 82.62$^{\pm0.00}$ & 14.57$^{\pm0.01}$ & 83.15$^{\pm0.00}$ & \textbf{20.48$^{\pm0.02}$} & \textbf{84.25$^{\pm0.00}$} \\
 hye  & 11.68$^{\pm0.01}$ & 83.64$^{\pm0.00}$ & 14.84$^{\pm0.01}$ & 86.49$^{\pm0.00}$ & \textbf{20.48$^{\pm0.02}$} & \textbf{87.76$^{\pm0.00}$} \\
 kat  & 11.86$^{\pm0.01}$ & 84.88$^{\pm0.00}$ & 12.61$^{\pm0.01}$ & 85.96$^{\pm0.00}$ & \textbf{20.12$^{\pm0.02}$} & \textbf{87.63$^{\pm0.00}$} \\
 khm  & 3.49$^{\pm0.00}$  & 79.13$^{\pm0.00}$ & 10.83$^{\pm0.01}$ & 84.96$^{\pm0.00}$ & \textbf{15.02$^{\pm0.01}$} & \textbf{85.91$^{\pm0.00}$} \\
 kmr  & 9.77$^{\pm0.01}$  & 82.07$^{\pm0.00}$ & 11.69$^{\pm0.01}$ & 84.18$^{\pm0.00}$ & \textbf{19.70$^{\pm0.02}$} & \textbf{85.88$^{\pm0.00}$} \\
 lao  & 6.41$^{\pm0.00}$  & 81.26$^{\pm0.00}$ & 13.25$^{\pm0.00}$ & 85.81$^{\pm0.00}$ & \textbf{23.52$^{\pm0.01}$} & \textbf{87.78$^{\pm0.00}$} \\
 mkd  & 12.00$^{\pm0.01}$ & 84.27$^{\pm0.00}$ & 13.15$^{\pm0.01}$ & 86.00$^{\pm0.00}$ & \textbf{16.35$^{\pm0.01}$} & \textbf{86.70$^{\pm0.00}$} \\
 mya  & 8.10$^{\pm0.01}$  & 82.15$^{\pm0.00}$ & 8.80$^{\pm0.00}$  & 84.88$^{\pm0.00}$ & \textbf{23.88$^{\pm0.02}$} & \textbf{87.49$^{\pm0.00}$} \\
 pus  & 14.83$^{\pm0.00}$ & 82.33$^{\pm0.00}$ & 17.65$^{\pm0.00}$ & 85.37$^{\pm0.00}$ & \textbf{25.03$^{\pm0.01}$} & \textbf{87.11$^{\pm0.00}$} \\
 sna  & 8.46$^{\pm0.01}$  & 81.66$^{\pm0.00}$ & 9.66$^{\pm0.01}$  & 81.98$^{\pm0.00}$ & \textbf{16.72$^{\pm0.02}$} & \textbf{83.48$^{\pm0.00}$} \\
 som  & 13.44$^{\pm0.01}$ & 83.47$^{\pm0.00}$ & 15.61$^{\pm0.02}$ & 85.19$^{\pm0.01}$ & \textbf{23.83$^{\pm0.03}$} & \textbf{86.71$^{\pm0.01}$} \\ \bottomrule
\end{tabular}
\caption{Results for Translate-Summarize-Translate Pipeline approach.}
\label{tab:tst-results-st-err}
\end{table*}

\section{Tokenizers}
\label{sec:tokenizers}
For the all experiments in this work we used the mT5 tokenizer to compute ROUGE scores. 
For computing novelty and length, we made use of some language specific tokenizers rather than rely on subword tokenization.
For Haitian Creole, Georgian, Macedonian, and both varieties of Kurdish, we used utoken\footnote{\url{https://github.com/uhermjakob/utoken}}. 
For Armenian, we used Stanza \citep{qi-etal-2020-stanza}.
and we used khmernltk \citep{hoang-khmer-nltk} for Khmer.
The tokenizers used in this work matter both for caluculating ROUGE scores and for determining the mean novelty score. For non-latin scripts, using the rouge package in huggingface's evaluate\footnote{\url{https://github.com/huggingface/evaluate}} can result in zero or near zero scores for non-latin script languages without explicitly supplying a tokenizer. 

\section{Analysis}
\label{sec:analysis}

\begin{table}[h!]
\centering 
\small
\begin{tabular}{@{}lrrrrr@{}}
\toprule
Lang. & \multicolumn{1}{l}{LR} & \multicolumn{1}{l}{Base-} & \multicolumn{1}{l}{Extract-} & \multicolumn{1}{l}{Self-} & \multicolumn{1}{l}{Back-} \\ 
 & \multicolumn{1}{l}{Sum} & \multicolumn{1}{l}{line} & \multicolumn{1}{l}{Train} & \multicolumn{1}{l}{Train} & \multicolumn{1}{l}{Sum} \\ 
\midrule
ckb & 63.94 & 9.16 & 11.15 & 5.17 & 5.72 \\
hat & 50.46 & 12.74 & 9.64 & 5.58 & 6.53 \\
hye & 69.08 & 17.78 & 21.68 & 16.95 & 97.33 \\
kat & 44.02 & 32.14 & 11.04 & 4.22 & 6.56 \\
khm & 25.81 & 67.44 & 68.92 & 68.48 & 66.78 \\
kmr & 41.26 & 43.38 & 10.77 & 8.90 & 4.06 \\
mkd & 66.74 & 12.99 & 18.70 & 11.87 & 12.22 \\ \bottomrule
\end{tabular}
\caption{Mean novelty scores using bigrams.}
\label{tab:bigram-novelty}
\end{table}

We conducted further analysis of generated summaries using bigrams to compute mean novelty and also include the mean length of summaries. 
We include them here due to space constraints in the paper.
Table \ref{tab:bigram-novelty} shows the mean novelty scores for summaries computed using bigrams.

\section{LLM Prompts}
\label{sec:prompts}
For LLM experiments we use the following prompts. 
For Aya-101, we use 

``\texttt{Write a summary for the following article in <<LANGUAGE>>. \textbackslash n <<ARTICLE\_TEXT>>''},

where <<ARTICLE\_TEXT>> is replaced with the text of the article to be summarized and <<LANGUAGE>> is replaced with the desired language. 
For Mixtral and Llama3, we used: 

\texttt{``Write a summary for the following article in <<LANGUAGE>>. Write the summary in <<LANGUAGE>>. 
Do not provide a translation or explain anything.
Only provide the summary, do not provide any other information except for the summary in <<LANGUAGE>>. 
Summarize this article:\textbackslash n<<ARTICLE\_TEXT>>
''}.

\subsection{Larger LLMs}
Summarization Prompt: "Summarize the following article using only 2 sentences: <<ARTICLE\_TEXT>>"

Translation prompt: "Translate the following text into <<LANG>>:  <<ARTICLE\_TEXT>>"

\subsection{mPrometheus LLM as Judge}
\label{sec:llm-judge-prompt}
"criteria": "Does the model provide a summary of the input article text that has decent semantic coverage, factuality, is consistent with the original article, is informative, coherent, fluent, concise and written in the language the article is written in?",
        "score1\_description": "The model neglects to provide a summary or the summary is not in the intended language.",
        "score2\_description": "The model provides a response but it is not a good summary. The response is factually inaccurate, not very informative, or not very fluent.",
        "score3\_description": "The model provides a summary but it is lacking in some of the desired qualities of a good summary.",
        "score4\_description": "The model provides a reasonable summary of the input text that includes most of the desired qualities of a good summary.",
        "score5\_description": "The model provides an excellent summary that meets all of the requested criteria of a good summary."

\section{Example Model Output}
We show example augmentation data in Table \ref{tab:aug-example} and examples of LLMs generating English output in Table \ref{tab:llm-eng}.

\label{sec:example-out}
\begin{table*}[th!]
\centering
\small
\resizebox{\textwidth}{!}{%
\begin{tabular}{p{0.1\linewidth} p{0.3\linewidth} p{0.3\linewidth} p{0.3\linewidth}}
\toprule
         & Extractive                                                                                                                                                                                                                                                                                                                                                                                                                                                                                                                                                                                                                                                                                                                                                                                                                                                & Self-Train                                                                                                                                                                                                                                                                                                                                                                                                                                                                                                                                                                                                                                                                                                                                                                                                                                                & BackSum                                                                                                                                                                                                                                                                                                                                                                                                                                                                                                                                                                                                                                                                                                                                                                                                                                                                                                                                                                                                                                                                                                                                                                       \\
         \midrule
Summary  & Ji girîngtirîn pêşangehên wênekêşiya takekesî ku ji aliyê Moîn Haşemînesab ve hatine organîzekirin, em dikarin behsa pêşangeha Wêne û Muzîk "Notên Bêdawî" û pêşangeha wêne û muzîkê "Bîst û Yek" bikin...                                                                                                                                                                                                                                                                                                                                                                                                                                                     & Mêvanê vê xelekê ji bernama Deng û Reng, Moîn Haşimî Neseb e                                                                                                                                                                                                                                                                                                                                                                                                                                                                                                                                                                                                                                                                                                                                                                                              & Ji girîngtirîn pêşangehên wênekêşiya takekesî ku ji aliyê Moîn Haşemînesab ve hatine organîzekirin, em dikarin behsa pêşangeha Wêne û Muzîk "Notên Bêdawî" û pêşangeha wêne û muzîkê "Bîst û Yek" bikin...                                                                                                                                                                                                                                                                                                                                                                                                                                                                                                                                                                                                                                                                                                                                               \\                         \midrule  
Document & Jînenîgarî. Moîn Haşimî Neseb bi çêkirina kurtefîlma "Dîktasyon" û beşdarbûna di çendîn festîvalên navneteweyî yên sala 2021ê de, xelata baştirîn kurtefîlma Festîvala New Yorkê wergirt. Fîlm di Festîvala Fîlman a Glasgowê (TMFF)... & Jînenîgarî. Moîn Haşimî Neseb bi çêkirina kurtefîlma "Dîktasyon" û beşdarbûna di çendîn festîvalên navneteweyî yên sala 2021ê de, xelata baştirîn kurtefîlma Festîvala New Yorkê wergirt. Fîlm di Festîvala Fîlman a Glasgowê (TMFF)...& Roja Pêncşemê li bajarê New Yorkê, ji aliyê Moîn Haşemînesab ve hatine organîzekirin, em dikarin behsa pêşangeha Wêne û Muzîk "Notên Bêdawî"... \\
\bottomrule
\end{tabular}
}
\caption{Examples of Extractive, Self-Training, and Back-Summarization approaches to synthetic data creation using an example from Kurmanji Kurdish.}
\label{tab:aug-example}
\end{table*}

\begin{table*}[ht]
\centering
\small
\resizebox{\textwidth}{!}{%
\begin{tabular}{p{0.10\linewidth} p{0.8\linewidth}}
\toprule
Model   & Output                                                                                                                                                                                                                                                                                                                                                                                                                                                                                                                                                                                                                                                                                                                                                                                      \\ \midrule
Mixtral & I'm sorry, I'm having difficulty understanding the text you provided. It appears to be written in Lao language and contains some unusual characters. Could you please rephrase or translate the question into English so I can better understand and provide an accurate response?                                                                                                                                                                                                                                                                                                                                                                                                                                                                                                           \\
\midrule
Mixtral & Overall, Ambassador Goldberg's visit to Luang Prabang reflects the U.S. government's ongoing commitment to supporting sustainable development, cultural preservation, and consular services in Laos, while also fostering stronger bilateral relations between the two countries.                                                                                                                                                                                                                                                                                                                                                                                                                                                                                                           \\
\midrule
Llama3  & I'd be happy to help you with that! However, I need the article text to write a summary in Lao. Please provide the article text, and I'll do my best to summarize it for you in Lao.                                                                                                                                                                                                                                                                                                                                                                                                                                                                                                                                                                                                        \\
\midrule
Llama3  & The Hua Seng Hung Company is one of the companies that have received an investment from the United States. This company has a lot of potential and it's expected to grow rapidly. The company is involved in many fields such as real estate, finance, and technology...
\\ \bottomrule
\end{tabular}
}
\caption{Examples of LLM English output when prompted to summarize non-English news articles.}
\label{tab:llm-eng}
\end{table*}

\section{Novelty and Length}
\label{sec:novelty-length}
\textbf{How extractive or abstractive are the multilingual fine-tuned MT5 summaries?}
While models trained on synthetic data have an advantage in ROUGE score over the baselines trained on only the human written summaries, it is possible that summaries produced by these models are still lacking in certain ways despite having higher scores. 
In particular, models trained on Extract-Train or Back-Sum data are trained on summaries generated from extractive models. 
One concern could be that these models only learn to copy material from the text rather than synthesizing a novel summary. 
We further probe this issue by computing mean novelty scores for each summary. 
This score is the percentage of tokens that do not appear in the article text.
We compute this novelty score using tokenizers described in Appendix \ref{sec:tokenizers}.

As seen in Table \ref{tab:novelty-indiv}, the test set reference summaries have somewhat high novelty. 
Each individual model generally has lower mean novelty than the test set. 
We may have expected model trained on extractive summaries to be generally less novel than those trained on self-training; however this does not appear to be the case.
This also shows a hint at why Armenian has low ROUGE scores for the back-sum approach. 
With such a high mean novelty score, there is evidence the model is generating a larger number of irrelevant words.

We show the novelty scores for the multilingual models in Table \ref{tab:multilingual-novelty}.
Similarly we see lower mean novelty from the multilingual models with augmented training data than the reference summaries. 

\begin{table*}[h!]
\centering
\small
\resizebox{.8\textwidth}{!}{%
\begin{tabular}{@{}lrrrrrrrr@{}}
\toprule
      & \multicolumn{4}{c}{Novelty}   &  \multicolumn{4}{c}{Length}     \\ \cmidrule(lr){2-5} \cmidrule(lr){6-9}
Lang. & \multicolumn{1}{l}{Reference} & \multicolumn{1}{l}{Extractive} & \multicolumn{1}{l}{Self-Train} & \multicolumn{1}{l}{BackSum} & \multicolumn{1}{l}{Reference} & \multicolumn{1}{l}{Extractive} & \multicolumn{1}{l}{Self-Train} & \multicolumn{1}{l}{BackSum} \\
\midrule
amh   & 49.8                          & 7.2                            & 9.5                            & 12.7                        & 25.1                          & 16.6                           & 15.8                           & 15.5                        \\
ckb   & 38.9                          & 1.6                            & 1.6                            & 2.0                         & 23.3                          & 26.1                           & 25.0                           & 26.1                        \\
hat   & 18.1                          & 1.0                            & 1.6                            & 2.6                         & 26.7                          & 27.4                           & 26.5                           & 25.8                        \\
hye   & 35.4                          & 2.9                            & 7.2                            & 4.0                         & 24.6                          & 18.5                           & 17.3                           & 18.5                        \\
kat   & 22.3                          & 2.3                            & 6.4                            & 3.4                         & 14.7                          & 15.8                           & 14.5                           & 16.0                        \\
khm   & 7.8                           & 9.3                            & 9.0                            & 8.8                         & 31.6                          & 64.8                           & 59.4                           & 60.6                        \\
kmr   & 16.3                          & 0.5                            & 25.0                           & 1.4                         & 20.2                          & 22.4                           & 16.1                           & 22.7                        \\
lao   & 22.5                          & 13.2                           & 12.3                           & 12.4                        & 28.4                          & 30.0                           & 29.6                           & 29.8                        \\
mkd   & 31.4                          & 2.3                            & 3.4                            & 3.0                         & 20.0                          & 20.3                           & 20.2                           & 20.8                        \\
mya   & 20.0                          & 8.4                            & 9.1                            & 9.1                         & 35.4                          & 33.7                           & 32.6                           & 33.1                        \\
pus   & 21.1                          & 3.1                            & 3.0                            & 3.7                         & 33.2                          & 25.2                           & 25.3                           & 25.1                        \\
sna   & 31.7                          & 1.6                            & 13.3                           & 1.9                         & 17.6                          & 18.4                           & 17.5                           & 18.2                        \\
som   & 26.0                          & 2.8                            & 4.4                            & 5.8                         & 24.7                          & 25.0                           & 23.0                           & 21.6                        \\ \bottomrule
\end{tabular}
}
\caption{Mean novelty scores and mean lengths of generated summaries by the multilingual models.}
\label{tab:multilingual-novelty}
\end{table*}

\begin{table}[h!]
\small
\centering
\begin{tabular}{@{}lrrrrr@{}}
\toprule
 & \multicolumn{1}{l}{Ref.} & \multicolumn{1}{l}{Base-} & \multicolumn{1}{l}{Extract-} & \multicolumn{1}{l}{Self-} & \multicolumn{1}{l}{Back-} \\
 & \multicolumn{1}{l}{} & \multicolumn{1}{l}{line} & \multicolumn{1}{l}{Train} & \multicolumn{1}{l}{Train} & \multicolumn{1}{l}{Sum} \\\midrule
ckb & 38.9 & 3.0 & 4.6 & 1.7 & 2.0 \\
hat & 18.1 & 6.7 & 2.6 & 1.4 & 2.0 \\
hye & 35.4 & 5.7 & 6.8 & 4.5 & 66.0 \\
kat & 22.3 & 19.1 & 4.2 & 1.4 & 1.9 \\
khm & 7.8 & 6.6 & 7.6 & 6.9 & 6.2 \\
kmr & 16.3 & 15.8 & 3.9 & 4.1 & 1.2 \\
mkd & 31.4 & 4.7 & 6.4 & 3.7 & 3.9 \\ \bottomrule
\end{tabular}
\caption{Mean Novelty for summaries generated by individual models and the summaries of the test set (LR-Sum)}
\label{tab:novelty-indiv}
\end{table}

\begin{table}[h!]
\centering
\small
\begin{tabular}{@{}lrrrrr@{}}
\toprule
Lang. & \multicolumn{1}{l}{Ref.} & \multicolumn{1}{l}{Base-} & \multicolumn{1}{l}{Extract-} & \multicolumn{1}{l}{Self-} & \multicolumn{1}{l}{Back-} \\ 
 & \multicolumn{1}{l}{} & \multicolumn{1}{l}{line} & \multicolumn{1}{l}{Train} & \multicolumn{1}{l}{Train} & \multicolumn{1}{l}{Sum} \\ \midrule
ckb & 23.3 & 25.1 & 27.9 & 25.0 & 26.8 \\
hat & 26.7 & 20.9 & 31.4 & 26.9 & 29.4 \\
hye & 24.6 & 22.2 & 19.9 & 16.9 & 16.8 \\
kat & 14.7 & 17.9 & 16.7 & 14.7 & 15.4 \\
khm & 31.6 & 71.2 & 74.9 & 69.0 & 71.1 \\
kmr & 20.2 & 15.9 & 27.4 & 21.2 & 22.6 \\
mkd & 20.0 & 21.6 & 21.2 & 19.9 & 21.5 \\ \bottomrule
\end{tabular}
\caption{The mean lengths for summaries generated by individual models in terms of tokens.}
\label{tab:leng}
\end{table}

\section{Evaluation Figures}
\label{sec:eval-figures}
\begin{figure*}[htb]
\centering
\small
    \includegraphics[width=\linewidth]{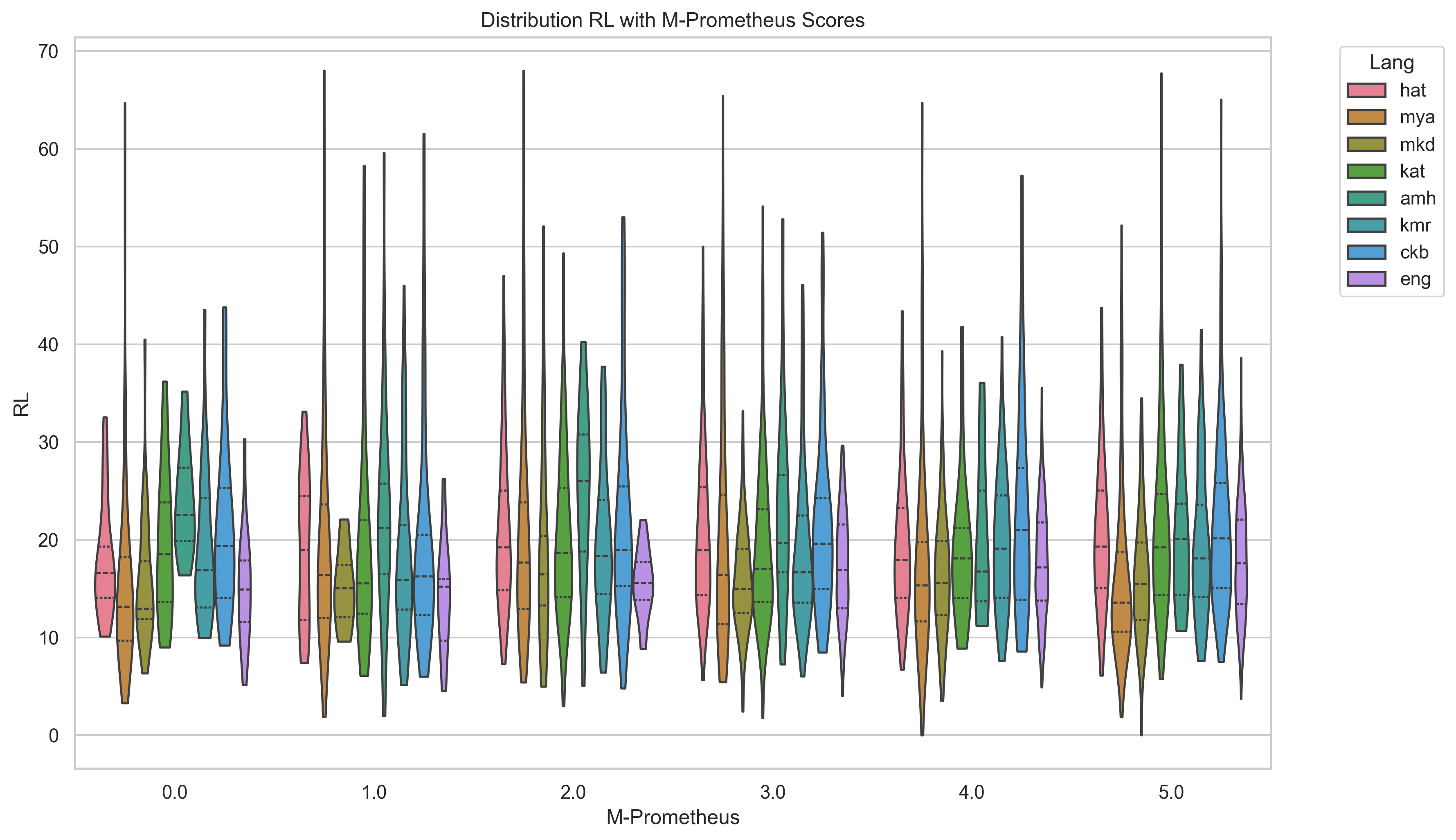}
    \caption{Distribution of summary scores for Llama 3.3 comparing scoring methods RL and M-Prometheus.}
    \label{fig:m-prometheus-RL}
\end{figure*}

\begin{figure*}[htb]
\centering
\small
    \includegraphics[width=\linewidth]{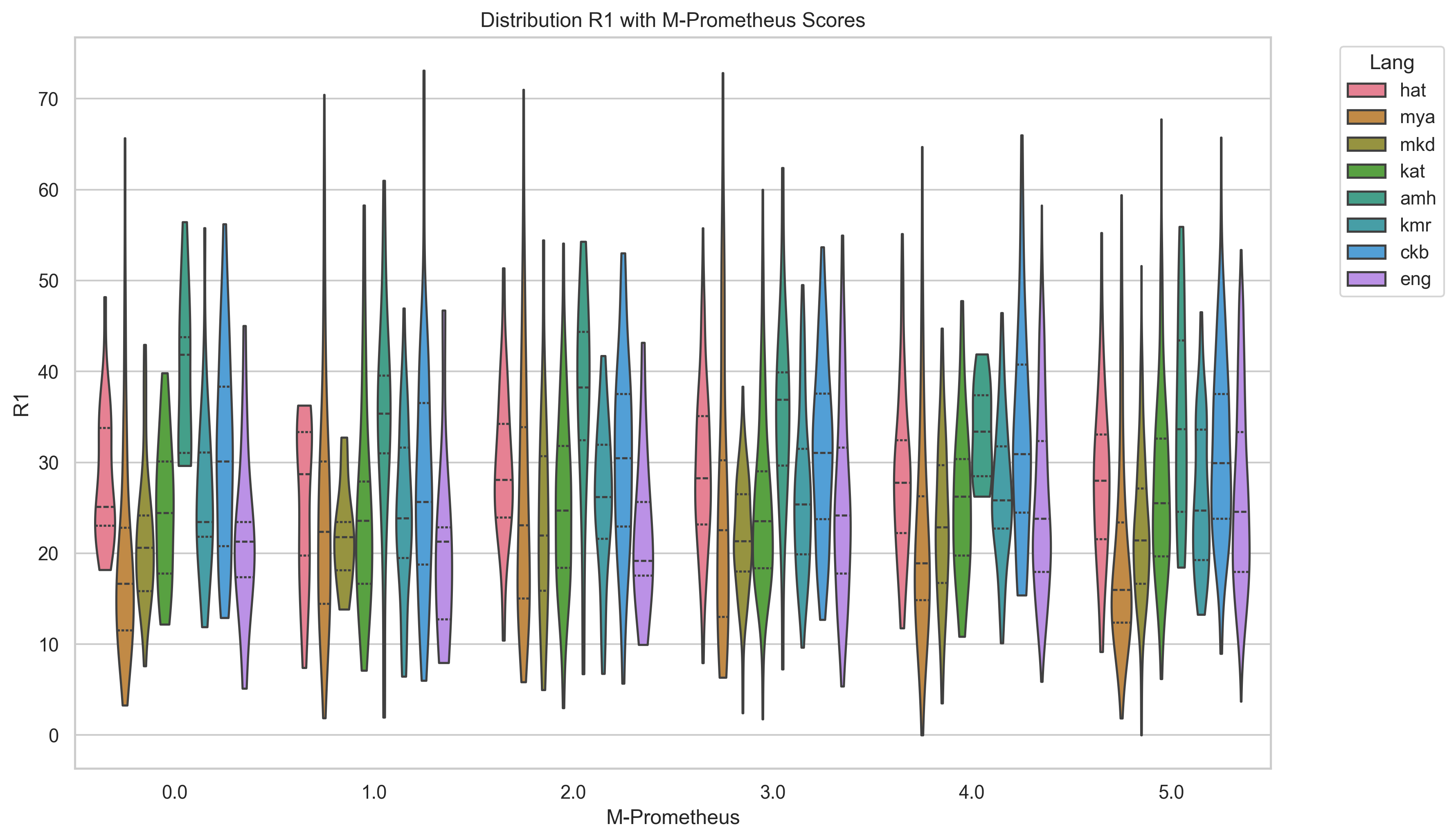}
    \caption{Distribution of summary scores for Llama 3.3 comparing scoring methods ROUGE-1 and M-Prometheus.}
    \label{fig:m-prometheus-R1}
\end{figure*}

\begin{figure*}[h!]
\centering
\small
    \includegraphics[width=\linewidth]{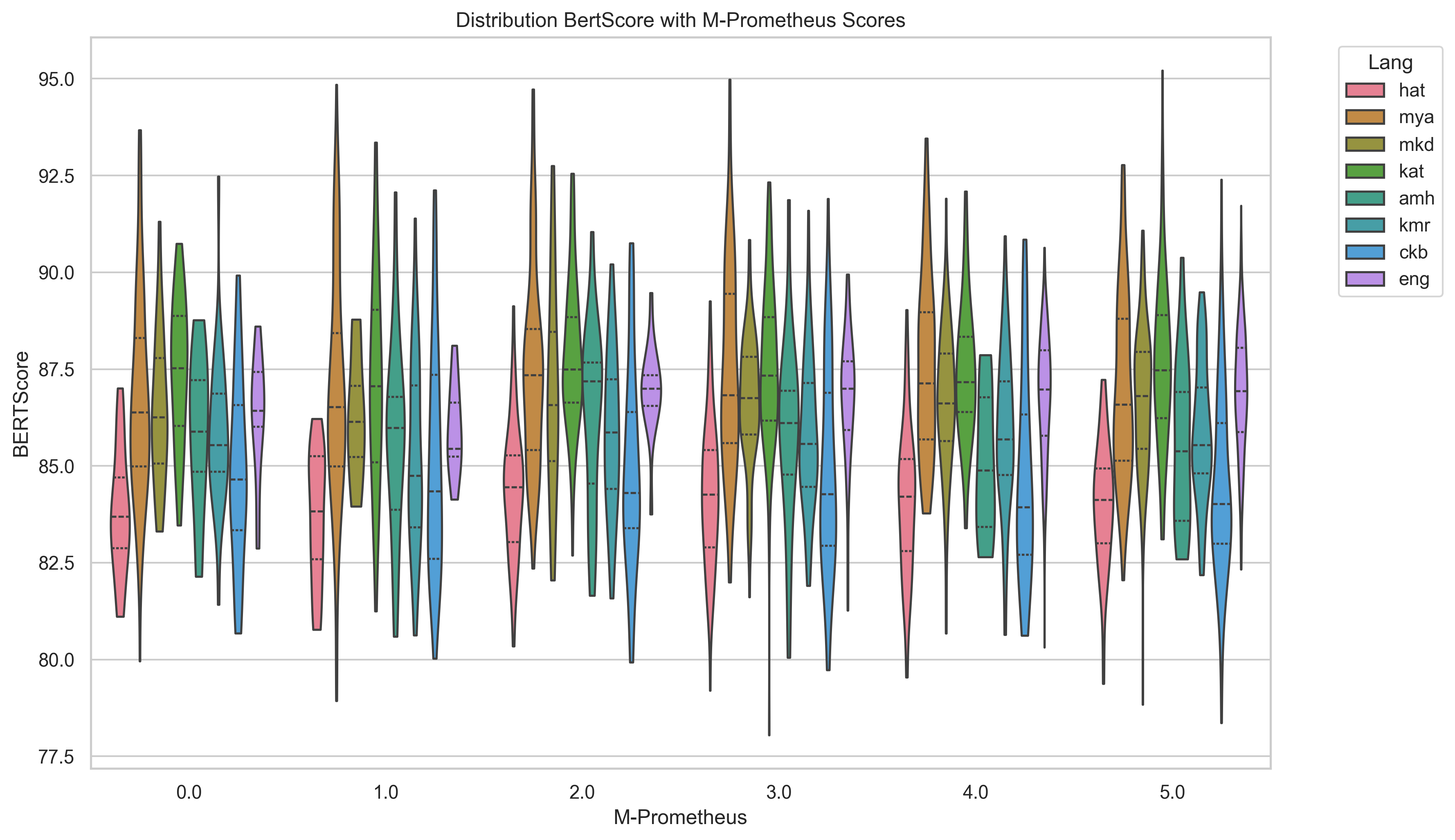}
    \caption{Distribution of summary scores for Llama 3.3 comparing scoring methods BERTScore and M-Prometheus.}
    \label{fig:m-prometheus-bertscore}
\end{figure*}

We present additional evaluation figures here.
Figure \ref{fig:m-prometheus-RL}, Figure \ref{fig:m-prometheus-R1}, and Figure \ref{fig:m-prometheus-bertscore} compare M-Prometheus scores with BERTScore and ROUGE scores. 

\section{Datasets}
\label{sec:app-datasets}
\begin{table*}[h!]
\centering 
\small
\resizebox{\textwidth}{!}{%
\begin{tabular}{@{}lllrrrrr@{}}
\toprule
Language & ISO & Lang.  & \multicolumn{1}{l}{Train} & \multicolumn{1}{l}{Train} & \multicolumn{1}{l}{Train} &  & \multicolumn{1}{r}{Wikipedia} \\
 & 639-3 & Family & LR-Sum &  XL-Sum & Combined & Wikipedia & Length Filtered \\ \midrule
Amharic          & amh       & Afro-Asiatic              & 0                                & 5,761                             & 5,761                               & 13,906                                 & 7,125                                            \\
Armenian         & hye       & Indo-European             & 920                              & 0                                & 920                                & 303,036                                & 287,288                                         \\
Burmese          & mya       & Sino-Tibetan              & 7,921                            & 4,569                            & 12,490                             & 109,310                                & 17,080                                          \\
Georgian         & kat       & Kartvelian                & 511                              & 0                                & 511                                & 169,602                                & 148,785                                         \\
Haitian Creole   & hat       & French Creole             & 452                              & 0                                & 452                                & 70,159                                 & 57,953                                          \\
Igbo             & ibo       & Niger-Congo (Volta-Niger) & 0                                & 4,183                            & 4,183                              & 22,908                                 & 20,496                                          \\
Khmer            & khm       & Austro-asiatic            & 3,888                            & 0                                & 3,888                              & 11,994                                 & 4,323                                           \\
Kurmanji Kurdish & kmr       & Indo-Iranian              & 791                              & 0                                & 791                                & 63,076                                 & 36,657                                          \\
Lao              & lao       & Kra-Dai                   & 11,964                           & 0                                & 11,964                             & 5,014                                  & 3,407                                           \\
Macedonian       & mkd       & Indo-European (Slavic)    & 1,223                            & 0                                & 1,223                              & 139,559                                & 122,754                                         \\
Oromo            & orm       & Afro-Asiatic (Cushitic)   & 0                                & 6,063                            & 6,063                              & 1,970                                  & 1,195                                           \\
Pashto           & pus       & Indo-Iranian              & 14,353                           & 16,854                           & 31,207                             & 20,529                                 & 15,308                                          \\
Scottish Gaelic  & gla       & Indo-European (Celtic)    & 1,313                            & 0                                & 1,313                              & 15,979                                 & 12,398                                          \\
Shona            & sna       & Niger-Congo (Bantu)       & 383                              & 0                                & 383                                & 11,621                                 & 9,963                                           \\
Sinhala          & sin       & Indo-Iranian              & 0                                & 3,249                            & 3,249                              & 23,065                                 & 16,782                                          \\
Somali           & som       & Afro-Asiatic (Cushitic)   & 0                                & 5,962                            & 5,962                              & 9,021                                  & 6,540                                           \\
Sorani Kurdish   & ckb       & Indo-Iranian              & 1,230                            & 0                                & 1,230                              & 52,024                                 & 35,098                                          \\
Yoruba           & yor       & Niger-Congo (Volta-Niger) & 6,350                            & 0                                & 6,350                              & 33,819                                 & 7,960                                           \\ \bottomrule
\end{tabular}
}
\caption{Language families and size of training data of LR-Sum and XL-Sum and available additional data from Wikipedia articles in number of documents before and after filtering for documents with more than 5 sentences.}
\label{tab:data}
\end{table*}
Table \ref{tab:data} shows counts of documents for each dataset.

\section{LLM Selection}
\label{sec:llm-selection}
Llama 3 8b Instruct \citep{dubey2024llama} and Mixtral 8x7b \citep{jiang2024mixtralexperts} are two open source LLMs which we use for benchmarking LLM performance in this work because of their reasonable performance on other benchmarks \citep{open-llm-leaderboard-v1,open-llm-leaderboard-v2} %
and ease of use for inference using Ollama.\footnote{\url{https://ollama.com}}
While Llama 3 and Mixtral perform reasonably on English, Aya-101  \citep{ustun-etal-2024-aya} is an LLM trained on much more multilingual data. 
We use these three LLMs off-the-shelf with simple prompts to compare our fine-tuned mT5 experiments.
For larger LLMs we use Gemma-3 27B \citep{gemma_2025}, Aya-Expanse 32B \citep{dang2024ayaexpansecombiningresearch}, and  Llama 3.3 70B. 
These models have reported reasonable performance and claim multilingual coverage.